%% file: Journal.tex
\documentclass[journal]{IEEEtran}

\input{preamble}
\begin{document}
\title{
Closed-Form, Provable, and Robust PCA via Leverage Statistics and Innovation Search}

\author{{Mostafa~Rahmani and Ping~Li}\\
Cognitive Computing Lab\\ Baidu Research\\
10900 NE 8th St. Bellevue, WA 98004, USA\\
\{rahmani.sut,\ pingli98\}@gmail.com
}
\markboth{}%
{Shell \MakeLowercase{\textit{et al.}}: Bare Demo of IEEEtran.cls for Journals}
\maketitle

\begin{abstract}
The  idea   of Innovation Search, which was initially proposed for data clustering, was recently used for outlier detection. 
In the application of Innovation Search for outlier detection, the directions of innovation were utilized to measure the innovation of the data points. We study the Innovation Values computed by the Innovation Search algorithm under a quadratic cost function and it is proved that Innovation Values with the new cost function are equivalent to Leverage Scores. This interesting connection is utilized to establish several theoretical guarantees for a Leverage Score based robust PCA method and to design a new robust PCA method.  The  theoretical results include performance guarantees with  different models for the distribution of  outliers and the  distribution of inliers. 
In addition, we demonstrate the robustness of the algorithms against the presence of noise.
 The  numerical and theoretical studies indicate that while the presented approach is fast and closed-form, it can outperform most of the existing algorithms.
\end{abstract}


\IEEEpeerreviewmaketitle

\section{Introduction}
Principal Component Analysis (PCA) has been extensively
used for linear dimensionality reduction.
While PCA is useful when the data has
low intrinsic dimension, its output is  sensitive to outliers
in the sense that the subspace found by  PCA can arbitrarily deviate from
the true underlying subspace even if a small portion of the data  is corrupted. In addition, locating the outlying components is of great interest in many applications. 
There are two different robust PCA problems corresponding to two different models for the data corruption. The first problem, known as low rank plus sparse matrix decomposition, assumes that a random subset of the elements of data are corrupted and the corrupted elements are not concentrated in any column/row of the data~\cite{lamport22,lamport1}. In the second problem,  a subset of columns of the data  \textcolor{black}{are affected by the data corruption}~\cite{lamport10,zhang2014novel,lerman2014fast,fischler1981random,li1985projection,choulakian2006l1,feng2012robust,mccoy2011two,hardt2013algorithms,zhang2016robust,you2017provable,markopoulos2014optimal}.
This paper focuses on the column-wise model, i.e., it is assumed that
data matrix $\bD \in \mathbb{R}^{M_1 \times M_2}$ can be expressed as
$$
\bD = ( [\bB \hspace{.2cm} \bA] ) \: \bT \:,
$$
where $\bA \in \mathbb{R}^{M_1 \times n_i}$, $\bB \in \mathbb{R}^{M_1 \times n_o}$,  $\bT$ is an unknown permutation matrix, and $[\bB \hspace{.2cm} \bA]$ represents the concatenation of matrices $\bA$ and $\bB$.
 The columns of $\bA$ lie in an $r$-dimensional subspace $\calU$. The columns of $\bB$ do not lie entirely in $\calU$, i.e., the $n_i$ columns of $\bA$ are the inliers and the $n_o$ columns of $\bB$ are the outliers.  
The output of a robust PCA method is an estimate for $\calU$.
If  $\calU$ is estimated accurately, the outliers can be  located by projecting the data points on the complement of $\calU$.

\subsection{Summary of Contributions}
Most of the existing robust PCA algorithms require a large number of iterations each with high computational complexity and most of them are not supported with thorough performance guarantees. We present  closed-form and provable robust PCA methods which  mostly outperform the existing methods.  The main contributions can be summarized as follows. 
\\
$\bullet$ It is proved that  Innovation Value introduced in~\cite{rahmani2019outlier2} is equivalent to Leverage Score if a quadratic cost function is used to find the optimal directions instead of the $\ell_1$-norm based cost function. This interesting connection is used to establish several  theoretical guarantees.
\\
$\bullet$ Inspired by the explanation of Leverage Scores with  Innovation Search, a new robust PCA method, which uses a symmetric measure of similarity, is presented. The presented closed-form methods mostly outperform the existing methods while they only include one singular value decomposition plus one matrix multiplication. 
\\
$\bullet$ Theoretical performance guarantees under several different models for the distribution of the outliers and inliers are presented. Furthermore, the robustness to the presence of noise is studied and it is shown that the  algorithm can provably distinguish the outliers when the data is noisy.



\subsection{Notation} 
Given a matrix $\bA$, $\| \bA \|$ denotes its spectral norm, $\| \bA \|_F$ denotes its Frobenius norm,  and $\bA^T$ is the transpose of $\bA$. For a vector $\ba$, $\| \ba \|_p$ denotes its $\ell_p$-norm and $\ba(i)$ its $i^{\text{th}}$ element.  For a matrix $\bA$, $\ba_i$ denotes its $i^{\text{th}}$ column and  $\| \bA \|_{1,2} = \sum_{i} \| \ba_i \|_2$.
$\mathbb{S}^{M_1 - 1}$ indicates the unit $\ell_2$-norm sphere in $\mathbb{R}^{M_1}$. Matrix $\bD = \bU^{'} \Sigma \bV$ where $\bU^{'} \in \mathbb{R}^{M_1 \times r_d}$ is the matrix of left singular vectors, $\Sigma \in \mathbb{R}^{r_d \times r_d}$ is a diagonal matrix whose diagonal values are equal to the non-zero singular values of $\bD$,  the rows of $\bV \in \mathbb{R}^{r_d \times M_2}$ are equal to the right singular vectors, and $r_d$ is the rank of $\bD$. The orthonormal matrix $\bU \in \mathbb{R}^{M_1 \times r}$ is defined as a basis for $\calU$. 
Note that $\bU^{'}$ is a basis for the entire data, $\bU$ is a basis for the inliers, $r_d$ is the rank of $\bD$, and $r$ is the rank of $\bA$. The subspace $\calU^{\perp}$ is defined as the complement of $\calU$.

\section{Related Work}
\label{sec:related_work}
 Robust PCA is a well-known problem and many approaches were developed for this problem. In this section, we briefly review some of the previous works on robust PCA.
Some of the earliest approaches to robust PCA are based on 
robust estimation of the data covariance matrix, such as the minimum covariance determinant, the minimum volume ellipsoid, and the Stahel-Donoho estimator~\cite{lamport47,feng2012robust,xu2010principal}.
However, these methods mostly compute a full SVD
or eigenvalue decomposition in each iteration and their performance greatly degrades when $\frac{n_i}{n_o} < 0.5$. In addition, they lack performance guarantees with structured outliers. 
Another approach is to replace the Frobenius Norm in the cost function of PCA  with  $\ell_1$-norm~\cite{lamport18,lamport29}, as $\ell_1$-norm was shown to be robust to the presence of the outliers~\cite{decod,lamport29}. In order to leverage the column-wise structure of the outliers,  
the authors of~\cite{lamport21} replaced the $\ell_1$-norm minimization problem used in~\cite{lamport29}  with an $\ell_{1,2}$-norm minimization problem. In~\cite{lerman2015robustnn} and~\cite{zhang2014novel}, the optimization problem used in~\cite{lamport21} was relaxed to two different convex optimization problems. The authors of~\cite{lerman2015robustnn,zhang2014novel}  provided sufficient conditions \textcolor{black}{under which} the optimal points of the convex optimization problems proposed in~\cite{lerman2015robustnn,zhang2014novel} are guaranteed to yield an exact basis for  $\calU$ . 
The approach presented in~\cite{tsakiris2015dual} focused on the scenario in which the data is predominantly unstructured outliers and the number of outliers is larger than $M_1$.
In~\cite{tsakiris2015dual}, it is essential to assume that the outliers are randomly distributed on $\mathbb{S}^{M_1 - 1}$ and the inliers are distributed randomly on the intersection of $\mathbb{S}^{M_1 - 1}$ and $\calU$. 
The outlier detection method proposed in~\cite{soltanolkotabi2012geometric} assumes that the outliers are randomly distributed on $\mathbb{S}^{M_1 - 1}$ and a small number of them are not linearly dependent which means that~\cite{soltanolkotabi2012geometric} is not able to detect the linearly dependent outliers and the outliers which are close to each other. Another approach is based on decomposing the given $\bD$ into a low rank matrix and a column sparse matrix where the column sparse matrix models the presence of the outliers \cite{cherapanamjeri2017thresholding,lamport10}. However, this approach requires the number of outliers to be significantly smaller than the number of the outliers and the solver algorithms that are used to decompose the data need to compute a SVD of the data in each iteration.

The main shortcomings of the previous methods are sensitivity to structured outliers and the lack of comprehensive theoretical guarantees. The Coherence Pursuit method, proposed in~\cite{rahmani2017coherence}, was shown (theoretically and numerically) to be robust to different types of outliers. However, Coherence Pursuit can miss outliers which carry weak innovation with respect to the inliers. The iSearch algorithm, proposed in~\cite{rahmani2019outlier2}, was shown to notably outperform Coherence Pursuit in detecting outliers with weak innovation . Similar to Coherence Pursuit, the robustness of iSearch against different types of outliers were supported with several theoretical guarantees. However, in contrast to Coherence Pursuit which is a closed-form method, iSearch needs to run an iterative and computationally expensive solver to find the directions of innovation.
\textit{This paper presents  robust PCA methods which have the advantages of both (CoP and iSearch) algorithms: while they are closed-form algorithms, their ability in distinguishing outliers are on a par with iSearch. 
In the rest of this section, Coherence Pursuit and iSearch are reviewed in more details.}

\vspace{0.05in}
\noindent
\textbf{Coherence Pursuit (CoP):}
CoP  ~\cite{rahmani2017coherence} assigns a value, termed Coherence Value, to each data point and $\calU$ is recovered using the span of the data points with the highest Coherence Values. The Coherence Value corresponding to data point $\bd_i$ represents the similarity between $\bd_i$ and the rest of the data points. CoP uses inner-product to measure the similarity between data points and it distinguishes the outliers based on the fact that an inlier bears more resemblance to the rest of the data than an outlier. 

\vspace{0.05in}
\noindent
\textbf{Innovation Search (iSearch)~\cite{rahmani2019outlier2}:}
Innovation Pursuit was initially proposed as a data clustering algorithm~\cite{rahmani2017subspacedi,rahmani2015innovation}.  The authors of~\cite{rahmani2017subspacedi,rahmani2015innovation} showed that Innovation Pursuit can notably outperform the self-representation based clustering methods (e.g. Sparse Subspace Clustering~\cite{lamport7}) specifically when the clusters are close to each other. 
Innovation Pursuit computes an optimal direction corresponding to each data point $\bd_i$ which can be written as the optimal point of 
\begin{eqnarray}
\underset{ \bc}{\min} \: \:  \| \bc^T \bD \|_1 \quad \text{subject to} \quad \bc^T \bd_i = 1 \:.
\label{eq:el1_innov}
\end{eqnarray}
If $\bC^{*} \in \mathbb{R}^{M_1 \times M_2}$ contains all the optimal directions, Innovation Pursuit builds the adjacency matrix as $\bQ + \bQ^T  $ where $\bQ = |\bD^T \bC^{*}|$. In~\cite{rahmani2019outlier2}, it was shown that the optimal directions  can be utilized for outlier detection too. The approach proposed in~\cite{rahmani2019outlier2}, termed iSearch, assigns an Innovation Value to each data point and it distinguishes the outliers as the data points with the higher Innovation Values. The Innovation Value assigned to $\bd_i$ is computed as 
$$
\frac{1}{\| \bD^T \bc_i^{*} \|_1}
$$
where $\bc_i^{*}$ is the optimal point of (\ref{eq:el1_innov}).  iSearch needs to run an iterative solver to find the optimal directions. In contrast, the  methods presented in this paper are closed-form and they can be hundreds of time faster. 

\vspace{0.05in}
\noindent
\textbf{Leverage Statistics:}
In regression, Leverage Scores are defined as the diagonal values  of the hat matrix $\bX^T (\bX \bX^T)^{-1} \bX $ where $\bX \in \mathbb{R}^{m_1 \times m_2}$ is the design matrix~\cite{everitt2002cambridge}. Assuming that the rank of $\bX$ is equal to $m_1$, then the $i^{\text{th}}$ leverage score is equal to $\| {\bv_x}_i \|_2^2$ where the rows of $\bV_x \in \mathbb{R}^{m_1 \times m_2}$ are equal to the right singular vectors of $\bX$ and ${\bv_x}_i$ is the $i^{\text{th}}$  column of $\bV_x$. 
Leverage   has been typically  used  in the regression framework and there are few  works which focused on  using it for the robust PCA problem~\cite{mejia2017pca,naes1989leverage}. For instance,
\cite{mejia2017pca} utilized leverage to reject the  outlying time points in an functional magnetic resonance images (fMRI) run.
However, there is not still an analysis and full understating of Leverage in the robust PCA setting. We show that  Innovation Value introduced in~\cite{rahmani2019outlier2} is equivalent to Leverage Score if a quadratic cost function is used to find the optimal directions. This interesting connection is used to establish several  theoretical guarantees and to design a new robust PCA method. 


\begin{algorithm}
        \caption{Asymmetric Normalized Coherence Pursuit (ANCP) for Robust PCA }
{
\textbf{Input.} The inputs are the data matrix $\bD \in \mathbb{R}^{M_1 \times M_2}$ and $r$ which is the dimension of the recovered subspace.

\smallbreak
\textbf{1.} Normalize the $\ell_2$-norm of the columns of $\bD$, i.e., set $\bd_i$ equal to $\bd_i / \| \bd_i \|_2$ for all $1 \le i \le M_2$.

\smallbreak
\textbf{2.} Rows of $\bV \in \mathbb{R}^{r_d \times M_2  }$ are equal to the first $r_d$ right singular vectors of $\bD$ where $r_d$ is the rank of $\bD$.

\smallbreak
\textbf{3.} Define $\bx \in \mathbb{R}^{M_2}$,  vector of Normalized Coherence Values, as  
\begin{eqnarray}
\bx(i) = \frac{1}{\| \bv_i \|_2^2} \: . 
\label{eq:ANCP}
\end{eqnarray}

\smallbreak
\textbf{4. } Construct matrix $\bY$ from the
columns of $\bD$ corresponding to the largest elements of $\bx$ such that they span an r-dimensional subspace.

\smallbreak
\textbf{ Output:} The column-space of $\bY$ is the identified subspace.
 }
\end{algorithm}

\section{Robust PCA with Leverage and Innovation Search}
\label{sec:proposed}
The table of Algorithm 1 and the table of Algorithm 2 show the presented methods along with the used definitions. Algorithm 1  utilizes Leverage Scores  to rank the data points and the data points with the minimum Leverage Scores are used to build a basis for $\calU$. 
In this section, we show the underlying connection between Algorithm 1 and iSearch  and 
the motivation for naming Algorithm 1 ``Asymmetric Normalized Coherence Pursuit'' is explained. 
In addition, we explain the motivations behind the design of Algorithm 2 based on the connection between Algorithm 1 and iSearch.  
Both algorithms are closed-form and they are  faster than most of the existing methods. The computation complexity of ANCP is $\calO( r_d M_1 M_2)$  and the computation complexity of SNCP is $\calO( r_d M_1 M_2 + r_d M_2^2)$.

The presented robust PCA algorithms use the data points corresponding to the largest  Normalized Coherence Values   to form the basis matrix $\bY$. If the inliers are distributed uniformly at random in $\calU$, then $r$ data points corresponding to the $r$ largest Normalized Coherence Values span $\calU$ with high probability. 
However, in real data, the inliers form some clustering structure and the algorithm should continue adding new columns to $\bY$ until the columns of  $\bY$ span an $r$-dimensional subspace. It means that we need to check the singular values of $\bY$  multiple times. Two techniques can be utilized to  \textcolor{black}{avoid these extra steps}~\cite{rahmani2019outlier2,rahmani2017coherence22f}. The first approach is based on leveraging side information that we mostly have about the population of the outliers.  In most of the applications, we can have an upper-bound on $n_o/M_2$ because outliers are mostly associated with rare events. If we know that the number of outliers is less than $y$ $\%$ of the data,  matrix $\bY$ can be constructed using $(1 - y)$ $\%$ of the data columns which are corresponding to the largest Normalized Coherence Values. The second technique is to use the adaptive column sampling method proposed in~\cite{rahmani2017coherence22f} which uses subspace projection to avoid sampling redundant columns.

\begin{algorithm}
        \caption{Symmetric Normalized Coherence Pursuit (SNCP) for Robust PCA}
{
\textbf{Input.} The inputs are the data matrix $\bD \in \mathbb{R}^{M_1 \times M_2}$ and $r$ which is the dimension of the recovered subspace.

\smallbreak
\textbf{1.}   Similar to Step 1 in Algorithm 1.

\smallbreak
\textbf{2.} Define $\bV \in \mathbb{R}^{r_d \times M_2  }$ as in Algorithm 1. 

\smallbreak
\textbf{3.} The vector of Normalized Coherence Values is defined as
$$\bx(i) = \sum_{j=1}^{M_2} \frac{({\bv_i}^T \bv_{j})^2}{ \|\bv_i\|_2^2 \|\bv_j\|_2^2 }\:.$$

\smallbreak
\textbf{4. } Construct matrix $\bY$ as in Algorithm 1. 

\smallbreak
\textbf{ Output:} The column-space of $\bY$ is the identified subspace.
 }
\end{algorithm}

\subsection{Explaining Leverage Score for Robust PCA Using Innovation Search}\label{sec:expl}

Algorithm 1 ranks the data points based on the inverse of their leverage scores. The following lemma shows that Leverage Score is directly related to Innovation Value. 
\begin{lemma}
Suppose rows of $\bV \in \mathbb{R}^{r_d \times M_2  }$ are equal to the first $r_d$ right singular vectors of $\bD$ where $r_d$ is the rank of $\bD$. Define $\bc_i^{*}$ as the optimal point of 
\begin{eqnarray}
\underset{ \bc}{\min} \: \:  \| \bc^T \bD \|_2 \quad \text{subject to} \quad \bc^T \bd_i = 1  \:.
\label{eq:el_2_inno}
\end{eqnarray}
Then,
$
\| \bD^T \bc_i^{*} \|_2^2 = \frac{1}{\| \bv_i \|_2^2} \:.
$

\label{lem:equi}
\end{lemma}
\noindent
Lemma~\ref{lem:equi} indicates that if a quadratic cost function  is used to compute the optimal directions in iSearch, Innovation Values are equivalent to Leverage Scores. Accordingly, we can use the idea of Innovation Search to explain the Leverage Score based robust PCA method. 
First suppose that  $\bd_i$ is an outlier which means that $\bd_i$ has a non-zero projection on $\calU^{\perp}$. Since most of the data points are inliers, the optimization problem utilizes the projection of $\bd$ in $\calU^{\perp}$ and finds the optimal direction near $\calU^{\perp}$ to minimize $\| \bA^T \bc_i\|_2^2$.
In sharp contrast, when $\bd_i$ is an inlier, the linear constraint strongly discourages the optimal direction to be close to $\calU^{\perp}$. Thus, $\| \bA^T \bc_i^{*}\|_2^2$ is notably larger when $\bd_i$ is an inlier comparing to $\| \bA^T \bc_i^{*}\|_2$  when $\bd_i$ is an outlier.
Accordingly, since $\frac{1}{\| \bv_i \|_2^2} = \| \bD^T \bc_i^{*}\|_2^2 = \|\bA^T \bc_i^{*} \|_2^2 + \|\bB^T \bc_i^{*} \|_2^2$, $1/\| \bv_i \|_2^2$
is much larger when $\bd_i$ is an inlier comparing to the same value when $\bd_i$ is an outlier because $\|\bA^T \bc_i^* \|_2^2$ is much larger when $\bd_i$ is an inlier.  

The following Lemma indicates that $\| \bD^T \bc_i^{*} \|_2^2$ can be written as the sum of the similarities between the columns of $\bV \in \mathbb{R}^{r_d \times M_2}$. 
\begin{lemma}
Define $\bc_i^{*}$ and $\bV$ as in Lemma~\ref{lem:equi}. Then,
\begin{eqnarray}
\| \bD^T \bc_i^{*} \|_2^2 = \sum_{j=1}^{M_2} \left(\frac{ \bv_i^T \bv_j }{ \|\bv_i\|_2 \|\bv_i\|_2 }\right)^2 \:. 
\label{eq:sum_asymmm}
\end{eqnarray}
\label{lm:secondlm} 
\end{lemma}
Thus,  Algorithm 1 is inherently similar to  CoP  but Algorithm 1 utilizes the coherency between the columns of  $\bV$. In other word, the functionality of Algorithm 1 is similar to that of a CoP algorithm which is applied to a data matrix whose non-zero singular values are normalized to 1.   This is the motivation  to name the  presented algorithms Normalized Coherence~Pursuit.

\subsection{Symmetric Normalized Coherence Pursuit}
The measure of similarity used in (\ref{eq:sum_asymmm}) is not symmetric. In other word, the similarity between $\bd_i$ and $\bd_j$ which is computed as $\left(\frac{ \bv_i^T \bv_j }{ \|\bv_i\|_2 \|\bv_i\|_2 }\right)^2 $ is not equal to the similarity between $\bd_j$ and $\bd_i$ which can be written as $\left(\frac{ \bv_i^T \bv_j }{ \|\bv_j\|_2 \|\bv_j\|_2 }\right)^2$. Accordingly, we modify the measure of similarity used in (\ref{eq:sum_asymmm}) into a symmetric measure of similarity. The Normalized Coherence Value corresponding to $\bd_i$ using the symmetric measure of similarity is defined as 
$
\bx(i) = \sum_{j=1}^{M_2} \left(\frac{ \bv_i^T \bv_j }{ \|\bv_i\|_2 \|\bv_j\|_2 }\right)^2 \:. 
$
Algorithm 2 uses the symmetric measure of similarity to compute the Normalized Coherence Values. The numerical experiments show that utilizing the symmetric function can notably improve the performance in most of the cases.  

\section{Theoretical Studies}
In this section, we present analytical performance guarantees for Normalize Coherence Pursuit under different models for the distribution of the outliers.  
The connection between Innovation Values and Leverage Scores is utilized to analyze the ANCP method and we leave the analysis of SNCP to future works. In the following subsections, the performance guarantees  with unstructured outliers, linearly dependant outliers, noisy inliers, and clustered outliers is provided. Moreover, in contrast to most of the previous methods whose guarantees are limited to randomly distributed inliers, Normalized Coherence Pursuit  is supported with theoretical guarantees even when the inliers are clustered. 
The following sections provide the theoretical results and each theorem is followed by a short  discussion
which highlights the important aspects of that theorem.




To simplify the exposition and notation, in the presented results,  it is assumed without loss of generality that $\bT$  is equal to the identity matrix, i.e, $\bD = [\bB \hspace{.2cm} \bA]$.
The subspace $\calU$ is recovered using the span of the data points with the largest Normalized Coherence Values. A sufficient condition which guarantees exact recovery of $\calU$ is that the minimum of the Normalized Coherence Values corresponding to the inliers is larger than the maximum of the Normalized Coherence Values corresponding to the outliers, i.e., 
\begin{eqnarray}
\begin{aligned}
&\min \left(  \{  \bx(i) \}_{i=n_o+1}^{M_2}  \right) >  \max \left(  \{ \bx(i) \}_{i=1}^{n_o}  \right) \:. 
\end{aligned}
\label{cond:main_cond}
\end{eqnarray}
\noindent
This
is not a necessary condition but it is easier to guarantee. 
In addition, we define $$\psi =  \max  \left( \left\{ \frac{1}{\| \bb_i^T \bR\|_2^2}  \right\}_{i=1}^{n_o} \right)$$ where $\bR$ is an orthonormal basis for $\calU^{\perp}$ and $\bb_i$ is the $i^{\text{th}}$ column of $\bB$. The parameter $\psi$ indicates how close the outliers are to $\calU$. 

\vspace{0.05in}
\noindent
\textbf{Proof Strategy:} Although the presented theorems consider different scenarios for the distribution of the inliers/outliers and different techniques are required to guarantee (\ref{cond:main_cond}) in each case, but a similar strategy is used in the proofs of all the results.  In contrast to Cop which analyzed $\{ |\bd_i^T \bd_j|\}_{i,j}$ based on the distribution of the data, it is not straightforward to directly bound $\{ |\bv_i^T \bv_j|\}_{i,j}$. In addition, in contrast to iSearch which leveraged the fact that the optimal direction of (\ref{eq:el1_innov}) is mostly orthogonal to $\calU$ when $\bd_i$ is an outlier, the optimal direction obtained by (\ref{eq:el_2_inno}) is not necessarily orthogonal to $\calU$ when $\bd_i$ is an outlier. In the proofs of the presented results, we utilized the geometry of the problem in which the optimal direction of (\ref{eq:el_2_inno}) is not  close to $\calU$ when $\bd_i$ is an outlier. Specifically, corresponding to outlier $\bd_i$, we define $\bd_i^{\perp} = \frac{\bR \bR^T \bd_i}{\| \bd_i^T \bR \|_2^2}$ and by definition $\bd_i^T \bd_i^{\perp} = 1$. According to the definition of $\bc_i^{*}$ as the optimal point of (\ref{eq:el_2_inno}) and according to Lemma~\ref{lem:equi}, $\frac{1}{\| \bv_i \|_2^2} \le \| \bD^T \bd_i^{\perp} \|_2^2 $. We utilized this inequality  to derive the sufficient conditions which prove that (\ref{cond:main_cond}) holds with high probability. The detailed proofs of all the results are provided in the appendix.

\subsection{Outliers Distributed on $\mathbb{S}^{M_1 -1}$}
\label{sec:uns}
In most of the previous works on the robust PCA problem, the performance of the outlier detection method is analyzed under the assumption that the outliers are randomly distributed on $\mathbb{S}^{M_1 - 1}$. This is a simple scenario because the outliers are unstructured and the projection of each outlier on $\calU$ is not strong with high probability given that $r$ is sufficiently small because when the outliers are randomly distributed as in Assumption \ref{assum_DistUni}, then $\mathbb{E} [\| \bU^T \bb \|_2^2] = \frac{r}{M}$ ( $\bb$ is an outlying data point). The following assumption specifies the presumed model for the distribution of the inliers/outliers. 

\begin{assumption}
The columns of $\bA$ are drawn  uniformly at random from $\calU \cap \mathbb{S}^{M_1 -1}$ and the columns of $\bB$ are drawn  uniformly at random from $\mathbb{S}^{M_1 - 1}$. 
\label{assum_DistUni}
\end{assumption}

The following theorem provides the sufficient conditions to guarantee the exact recovery of $\calU$.

\begin{theorem}
\label{theo:randomrandom}
Suppose $\bD$ follows Assumption 1.  If $\bx$ is defined as in  (\ref{eq:ANCP}) and
\begin{eqnarray} 
\begin{aligned}
& \frac{n_i}{r} - \max \left( \frac{4}{3} \log \frac{2r}{\delta} , \sqrt{4 \frac{n_i}{r} \log \frac{2 r}{\delta}} \right) > \frac{(\psi - 1)n_o}{M_1} + \\
& \quad \max \left( \frac{8}{3} \log \frac{2M_1}{\delta} , \sqrt{16 \frac{n_o}{M_1} \log \frac{2 M_1}{\delta}} \right) \:,
\end{aligned}
\label{eq:suff_1}
\end{eqnarray}
then (\ref{cond:main_cond}) holds and   $\calU$ is recovered exactly with probability at least $1 - 3 \delta$.
\label{theo:random}
\end{theorem}

Since the outliers are randomly distributed, the expected value of $\| \bb_i^T \bR \|_2^2$ is equal to $\frac{M_1- r}{M_1}$  which is nearly equal to~1 when $r/M_1$ is small~\cite{rahmani2017coherence,park2014greedy}.
Thus, Theorem~\ref{theo:random} roughly indicates that if ${n_i}/{r}$ is sufficiently larger than $n_o/M_1$, Normalized Coherence Values can successfully distinguish the outliers. It is important to note that $n_i$ is scaled with $r$ while $n_o$ is scaled with $M_1$. It means that if $r$ is sufficiently small and if the outliers are unstructured,   $\calU$ can be recovered exactly even if $n_o$ is much larger than $n_i$. 
In addition, one can observe that when the outliers are unstructured, the requirements of Normalized Coherence Pursuit are similar with those of CoP~\cite{rahmani2017coherence}.  In the next subsection, we observe a clear difference between their requirements when the outliers are close to $\calU$.

\subsection{Outliers in an Outlying Subspace}
\label{sec:outlier_sub}
Although Assumption~\ref{assum_DistUni} is a popular data model in the literature of robust PCA, it is not a realistic assumption in the practical scenarios. In practice, outliers can be structured and  they are not completely independent from each other as it is assumed in Assumption~\ref{assum_DistUni}.  
For instance, in anomaly event detection, the outlying video frames are highly correlated or in misclassified data points identification, they can belong to the same cluster~\cite{gitlin2018improving}. In this section, we study the robustness against linearly dependant outliers. The following assumption specifies the presumed model for the outliers.

\begin{assumption}
Define
subspace $\calU_o$ with dimension $r_o$ such that $\calU_o \notin \calU$ and $\calU \notin \calU_o$.
The columns of $\bA$ are randomly distributed on  $\calU \cap \mathbb{S}^{M_1 -1}$  and the columns of  $\bB$ are randomly distributed on   $\calU_o \cap \mathbb{S}^{M_1 - 1} $.
\label{asm:out}
\end{assumption}
The following theorem provides the sufficient condition to guarantee that (\ref{cond:main_cond}) holds. 
\begin{theorem}
\label{theo:Linearly_dependant}
Suppose  $\bD$ follows Assumption~\ref{asm:out}.  If $\bx$ is defined as in (\ref{eq:ANCP}) and
\begin{eqnarray*}
\begin{aligned}
&\frac{n_i}{r} - \max \left( \frac{4}{3} \log \frac{2r}{\delta} , \sqrt{4 \frac{n_i}{r} \log \frac{2 r}{\delta}} \right) >   \\
& \| \bU_{o}^T \bU^{\perp} \|\left( \frac{\psi n_o}{r_o} + \psi \max \left( \frac{4}{3} \log  \frac{2r_o}{\delta} , \sqrt{\frac{n_o}{r_o} \log \frac{2 r_o}{\delta}} \right)\right)
\end{aligned}
\end{eqnarray*}
then (\ref{cond:main_cond}) is satisfied  and $\calU$ is recovered exactly with probability at least $1 - 2 \delta$.
\end{theorem}

Theorem~\ref{theo:Linearly_dependant} roughly states that if $n_i/r$ is sufficiently larger than $n_o/r_o$, the exact recovery is guaranteed with high probability. If $r_o$ is comparable to $r$, then the number of inliers should be sufficiently larger than the number of outliers. This confirms our intuition about the outliers because if $r_o$ is comparable to $r$  and $n_o$ is also large, we cannot label the columns of $\bB$ as outliers. 
It is informative to compare the requirements of  Normalized Coherence Pursuit with that of CoP. The following theorem provides the sufficient conditions to guarantee that  CoP successfully distinguishes the outliers. 
\begin{theorem}
\label{theo:Linearly_dependant_CP}
Suppose $\bD$ follows  Assumption~\ref{asm:out}. If 
\begin{eqnarray*}
\begin{aligned}
& \frac{n_i}{r} - \max \left( \frac{4}{3} \log \frac{2r}{\delta} , \sqrt{4 \frac{n_i}{r} \log \frac{2 r}{\delta}} \right) > \\
 & \|\bU^T \bU_o \|^2 \left( \frac{n_i}{r} + \max \left( \frac{4}{3} \log \frac{2r}{\delta} , \sqrt{4 \frac{n_i}{r} \log \frac{2 r}{\delta}} \right) \right) + \\
 & \frac{n_o}{r_o} + \max \left( \frac{4}{3} \log \frac{2r_o}{\delta} , \sqrt{4 \frac{n_o}{r_o} \log \frac{2 r_o}{\delta}} \right) \: ,
\end{aligned}
\label{eq:cop_reqq}
\end{eqnarray*}
then the CoP method proposed in~\cite{rahmani2017coherence} recovers $\calU$ exactly with probability at least $1- 3\delta$.

\end{theorem}

One can observe that the requirement  of Theorem~\ref{theo:Linearly_dependant_CP} is much stronger than that of Theorem~\ref{theo:Linearly_dependant} because $n_i/r$ appears on the right hand side of the sufficient condition of Theorem~\ref{theo:Linearly_dependant_CP}.
 Theorem~\ref{theo:Linearly_dependant_CP} predicts that when $\calU_o$ is close $\calU$, the CoP Algorithm is more likely to fail. This is a correct prediction because when $\calU$ and $\calU_o$ are close, the inliers and the outliers are close to each other and their inner-product values are large. 
 In addition, by comparing the sufficient conditions of Normalized Coherence Pursuit with that of iSearch~\cite{rahmani2019outlier2} with linearly dependant outliers, we can observe that the nature of the sufficient conditions are similar. In the presented experiments, it is shown that Normalized Coherence Pursuit is on a par with iSearch in identifying outliers with weak innovation while it is a closed-from algorithm and its running time is much faster. 
\subsection{Noisy Inliers}
\label{sec:nooooise}
Although exact recovery of $\calU$ is not feasible when the inliers are noisy but the Normalized Coherence Values can distinguish the outliers even in the strong presence of noise. 
In this section , we present a theorem which guarantees that (\ref{cond:main_cond}) holds with high probability if $n_i/r$ and Signal to Noise Ratio (SNR) are sufficiently large. The following assumption specifies the presumed model.

\begin{assumption}
The matrix $\bD$ can be expressed as
$$
\bD =  [\bB \hspace{.2cm} \frac{1}{\sqrt{1+\sigma_n^2}}(\bA+\bE)] \: \bT \:.
$$
The matrix $\bE \in \mathbb{R}^{M_1 \times n_i}$ represents the presence of noise and it can be written as $\bE = \sigma_n \bN$ where the columns of $\bN \in \mathbb{R}^{M_1 \times n_i}$  are drawn  uniformly at random from $\mathbb{S}^{M_1 - 1}$ and $\sigma_n$ is a positive number which controls the power of the added noise.
\label{asm:noiss}
\end{assumption}

Before we state the theorem, let us define vectors $\{ \bt_i \}_{i=1}^{M_2}$ where $\bt_i = \Sigma \bv_i$ and the diagonal matrix $\Sigma$ contains the non-zero singular values of $\bD$. Note that $\bd_i = \bU^{'} \bt_i$ and $\| \bt_i \|_2 = \| \bd_i \|_2$. 
In addition,  define 
$
 t_{\min} = \min_i \left( \left\{ \frac{\| \Sigma^{-2} \bt_i \|_2}{\bt_i^T \Sigma^{-2} \bt_i}  \right\}_{i=n_o + 1}^{M_2} \right)$  and $ t_{\max} = \max_i \left( \left\{ \frac{\| \Sigma^{-2} \bt_i \|_2}{\bt_i^T \Sigma^{-2} \bt_i}  \right\}_{i=n_o + 1}^{M_2} \right) \:.
$

\begin{theorem}
\label{theo:noise}
Suppose $\bA$ and $\bB$ follow Assumption~\ref{assum_DistUni} and  $\bD$ is formed according to Assumption~\ref{asm:noiss}.
If  $\bx$ is defined as in  (\ref{eq:ANCP}) and
\begin{eqnarray*} 
\begin{aligned}
& \frac{(\sqrt{1 + \sigma_n^2}  - t_{\max} \sigma_n)^2}{1 + \sigma_n^2} \\
&  \Bigg( \frac{n_i}{r} -  \max \left( \frac{4}{3} \log \frac{2r}{\delta} , \sqrt{4 \frac{n_i}{r} \log \frac{2 r}{\delta}} \right) \Bigg) >  2 \sigma_n n_i t_{\max}^2 +   \\
& \psi \left( \frac{n_o}{M_1} + \max \left( \frac{4}{3} \log \frac{2 M_1}{\delta} , \sqrt{4 \frac{n_o}{M_1} \log \frac{2 M_1}{\delta}}  \right) \right) + \\
& \frac{\sigma_n^2 \psi}{1+\sigma_n^2 } \Bigg( \frac{n_i}{M_1} +  \max \left( \frac{4}{3} \log \frac{2 M_1}{\delta} , \sqrt{4 \frac{n_i}{M_1} \log \frac{2 M_1}{\delta}} \right) \Bigg)
\end{aligned}
 \end{eqnarray*} 
then (\ref{cond:main_cond}) holds  with probability at least $1 - 3 \delta$.
\end{theorem}

In this section, we considered the unstructured outliers whose number can be much larger than $M_1$ and $n_i$. Consider the challenging scenario that the unstructured outliers dominate the data, thus the values of all the singular values are close to each other which indicates that the values of $t_{\min}$ and $t_{\max}$ are close to one. Thus, the sufficient condition of Theorem~\ref{theo:noise}  roughly states that $\frac{(1- \sigma_n)^2}{1+ \sigma_n^2} \frac{n_i}{r}$ should be sufficiently larger than $n_i \sigma_n + \frac{n_o}{M_1}$. In practise, the algorithm works  better than what the sufficient condition implies 
because  the proof is based on considering  the worst case scenarios.  

\subsection{Clustered Outliers}
In this section, we consider a different structure for the outliers. It is assumed that the outliers form a cluster outside of the span of the inliers. Structured outliers are mostly associated with important rare events  such as malignant tissues~\cite{karrila2011comparison} or web attacks~\cite{kruegel2003anomaly}. The following assumption specifies the presumed model for $\bB$.

\begin{assumption}
Each column of $\bB$ is formed as $\bb_i = \frac{1}{\sqrt{1 + \eta^2}} ( \bq + \eta \mathbf{f}_i)$. The unit $\ell_2$-norm vector $\bq$ does not lie in $\calU$, $\{\mathbf{f}_i \}_{i=1}^{n_o}$ are drawn uniformly at
random from $\mathbb{S}^{M_1 - 1}$, and $\eta$ is a positive number.
\label{asm:clus}
\end{assumption}

\noindent
In Assumption~\ref{asm:clus}, the outliers form a cluster around vector $\bq$ which does not lie in $\calU$ and $\eta$ determines how close they are to each other. The following theorem provides the sufficient conditions to guarantee that Normalized Coherence Values distinguish the cluster of  outliers. 

\begin{theorem}
\label{theo:clustered_outliers}
Suppose that the distribution of  inliers follows Assumption ~\ref{assum_DistUni} and  the distribution of  outliers follows Assumption~\ref{asm:clus}. If $\bx$ is defined as  in (\ref{eq:ANCP}) and
\begin{eqnarray*} 
\begin{aligned}
& \frac{n_i}{r} - \max \left( \frac{4}{3} \log \frac{2r}{\delta} , \sqrt{4 \frac{n_i}{r} \log \frac{2 r}{\delta}} \right) > 
n_o \frac{\psi  \| \bq^T \bU^{\perp} \|_2^2}{1 + \eta^2} +  \\
& \frac{\psi \eta^2  n_o}{(1 + \eta^2 )M_1} +  \frac{\eta^2  \psi}{1 + \eta^2} \max \left( \frac{4}{3} \log \frac{2M_1}{\delta} , \sqrt{\frac{n_o}{M_1} \log \frac{2 M_1}{\delta}} \right) + \\
&\frac{ \eta \sqrt{\psi}}{1+ \eta^2} \| \bq^T \bU^{\perp} \|_2 \left( \frac{n_o}{\sqrt{M_1}} + 2\sqrt{n_o} + \sqrt{\frac{2 n_o \log \frac{1}{\delta}}{M_1 -1 }} \right) \:,
\end{aligned}
\end{eqnarray*} 
then (\ref{cond:main_cond}) is satisfied and   $\calU$ is recovered exactly with probability at least $1 - 4 \delta$.
\end{theorem}

In sharp contrast to Theorem~\ref{theo:randomrandom},  $n_o$ is not scaled with $M_1$. This means that when $\eta$ is small (the outliers are close to each other), $n_i/r$ should be sufficiently larger than $n_o$.  When  $\eta$ goes to infinity, the distribution of  outliers converges to the distribution of outliers  in Assumption~\ref{assum_DistUni} and one can observe that the sufficient condition in Theorem~\ref{theo:clustered_outliers} converges to the sufficient condition in Theorem~\ref{theo:randomrandom}.

\subsection{Outlier Detection in a Union of Subspaces}
In practice, the inliers are not randomly distributed in a subspace and they mostly form some structures. In this section, we assume that the inliers are clustered. It is assumed that the columns of $\bA$ form $m$ clusters and the data points in each cluster span a $d$-dimensional subspace. The following assumption provides the details. 
 
\begin{assumption}
The matrix of inliers can be written as $\bA = [\bA_1 \: ... \: \bA_m] \bT_A$ where $\bA_k \in \mathbb{R}^{M_1 \times {n_i}_k}$, $\sum_{k=1}^{m}  {n_i}_k = n_i$, and $\bT_A$ is an arbitrary permutation matrix.
The columns of $\bA_k$  are drawn uniformly at random from the
intersection of  subspace $\calU_k$ and $\mathbb{S}^{M_1-1}$ where $\calU_k$ is a $d$-dimensional subspace. In other word, the columns of $\bA$ lie in a union of subspaces $\{ \calU_k \}_{k=1}^m$ and $\left(\calU_1 \oplus \: ... \oplus \calU_m \right)= \calU$ where $\oplus$ denotes the direct sum operator.
\label{asm:union_of_sunb}
\end{assumption}

\noindent
The following theorem provides the sufficient conditions to guarantee that the computed Normalized Coherence Values satisfy (\ref{cond:main_cond}) with high probability.

\begin{theorem}
\label{theo:Linearly_dependant_inliers}
Suppose that the distribution of the inliers follows Assumption~\ref{asm:union_of_sunb} and the distribution of outliers follows Assumption~\ref{assum_DistUni}. If  $\bx$ is defined as in  (\ref{eq:ANCP}) and
\begin{eqnarray*}
\begin{aligned}
\vartheta \calA >  (\psi - 1)\frac{n_o}{M_1} +  2 \max \left( \frac{4}{3} \log \frac{2M_1}{\delta} , \sqrt{4 \frac{n_o}{M_1} \log \frac{2 M_1}{\delta}} \right)
\end{aligned}
\end{eqnarray*} 
where $ \vartheta = \underset{\ba \in \calU  \atop \| \ba \| = 1}{\inf} \sum_{k=1}^{m} \| \ba^T \bU_k \|_2^2$ and
$
 \calA = \min_i \Bigg\{ \frac{ {n_i}_k }{d} - \max \left( \frac{4}{3} \log \frac{2 m d}{\delta} , \sqrt{4 \frac{{n_i}_k}{d} \log \frac{2 m d}{\delta}} \right)  \Bigg\}_{i=1}^m, 
$
then (\ref{cond:main_cond}) is satisfied and  $\calU$ is recovered exactly with probability at least $1 - 3 \delta$.
\end{theorem}

Theorem~\ref{theo:Linearly_dependant_inliers} reveals an interesting property of the Normalized Coherence Values. 
According to the definition of $\calA$, $\calA$ is roughly equal to ${\min \{{n_i}_k \}_{k=1}^m}/{d}$. 
Thus, Theorem~\ref{theo:Linearly_dependant_inliers} states that when the inliers are clustered, the population of the cluster with the minimum population is the key factor. 
This property matches with our intuition about outlier detection because if there is a cluster with few number of data points, we could label them as outliers similar to the outliers modeled in Assumption~\ref{asm:out}.
The parameter $\vartheta = \underset{\ba \in \calU  \atop \| \ba \| = 1}{\inf} \sum_{k=1}^{m} \| \ba^T \bU_k \|_2^2$  shows  how well the inliers are diffused in $\calU$. Clearly, if the inliers are present in all or most of the directions inside $\calU$,
 a robust PCA algorithm is more likely to recover $\calU$ correctly. However, the presented methods do not require the inliers to occupy all the directions in $\calU$. The reason that
 $\vartheta$ appeared in the sufficient conditions is that the theorem guarantees  the performance in the worst case scenarios.

\section{Numerical Experiments}
In this section, SNCP and ANCP are compared with the existing robust PCA approaches, including FMS~\cite{lerman2014fast}, GMS~\cite{zhang2014novel}, CoP~\cite{rahmani2017coherence}, iSearch~\cite{rahmani2019outlier2}, and R1-PCA~\cite{lamport21},
and their robustness against different types of outliers   is examined with both real and synthetic data. 

\begin{remark}
In the presented theoretical results, it was assumed that $r_d$ is known. When the data is noisy, one can utilize any rank estimation algorithm and the performance of the algorithms is not sensitive to the chosen rank as long as $r_d$ is sufficiently larger than $r$.  In the presented experiments, we set $r_d$ equal to the number of singular values of $\bD$ which are greater than $s_1/20$ where $s_1$ is the first singular value of $\bD$. 
\end{remark}

\subsection{Comparing Different Scores}
\begin{figure}[t]
\begin{center}
\mbox{
\includegraphics[width=1.78in]{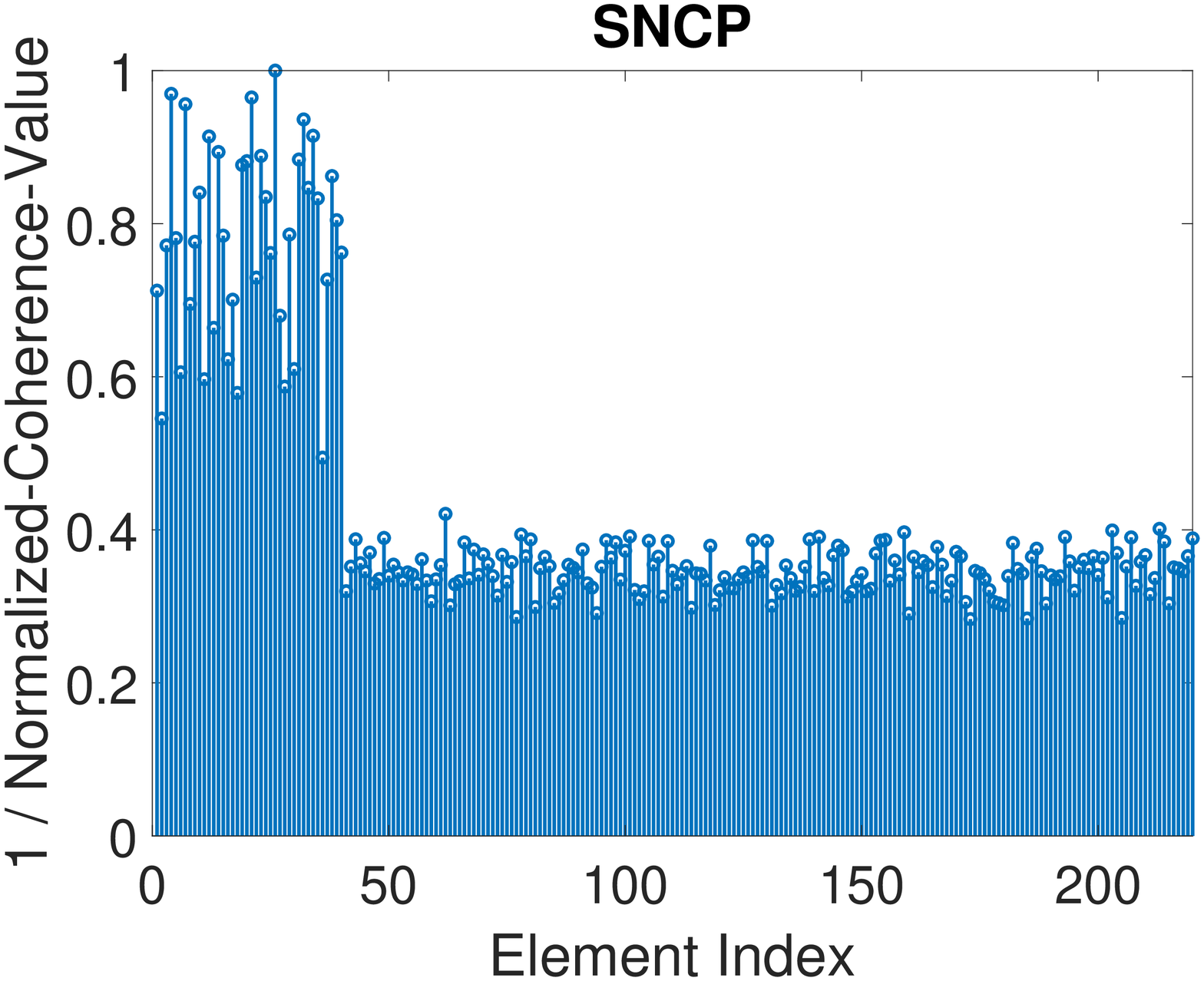}\hspace{-0.1in}
\includegraphics[width=1.78in]{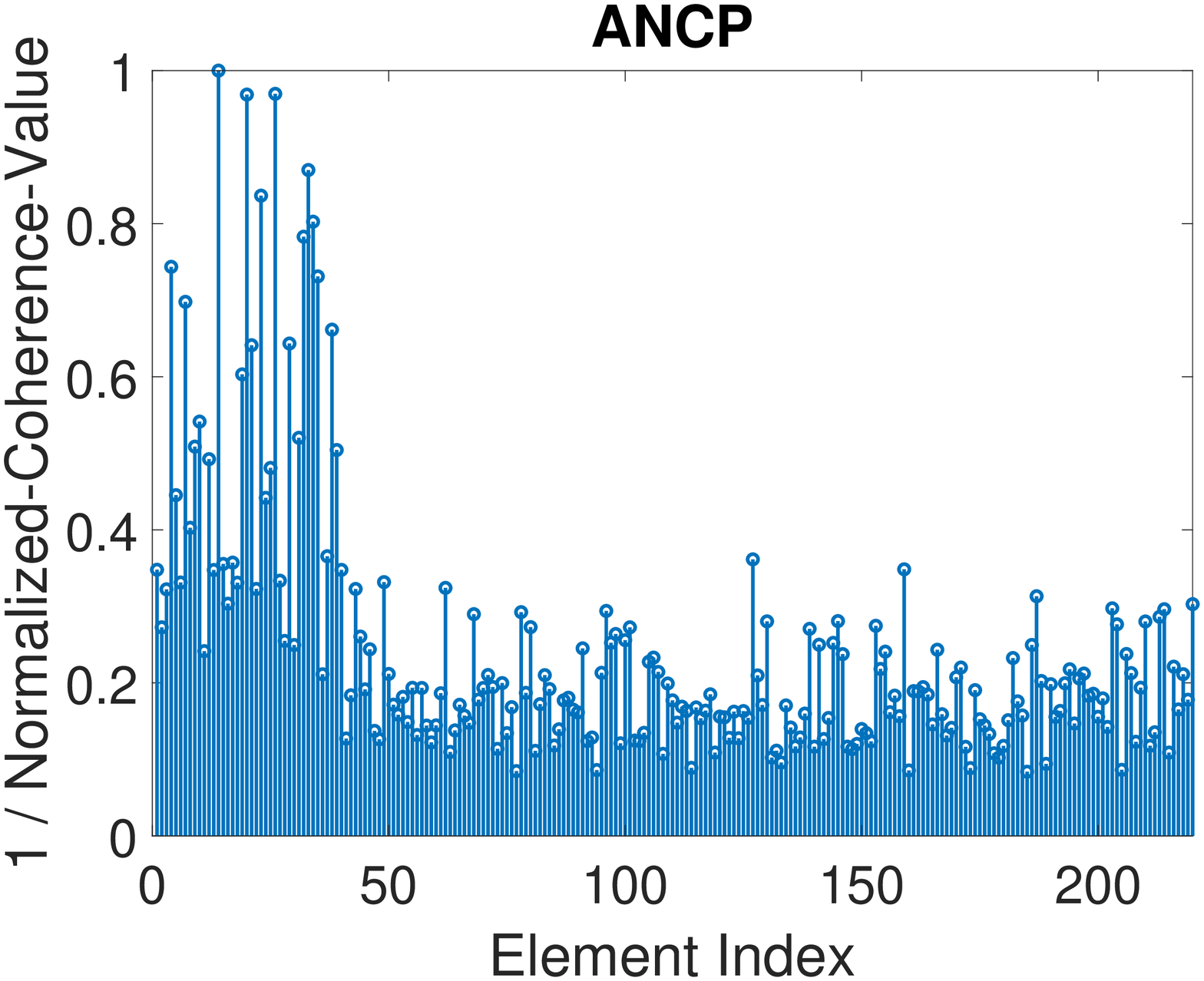}}
\mbox{
\includegraphics[width=1.78in]{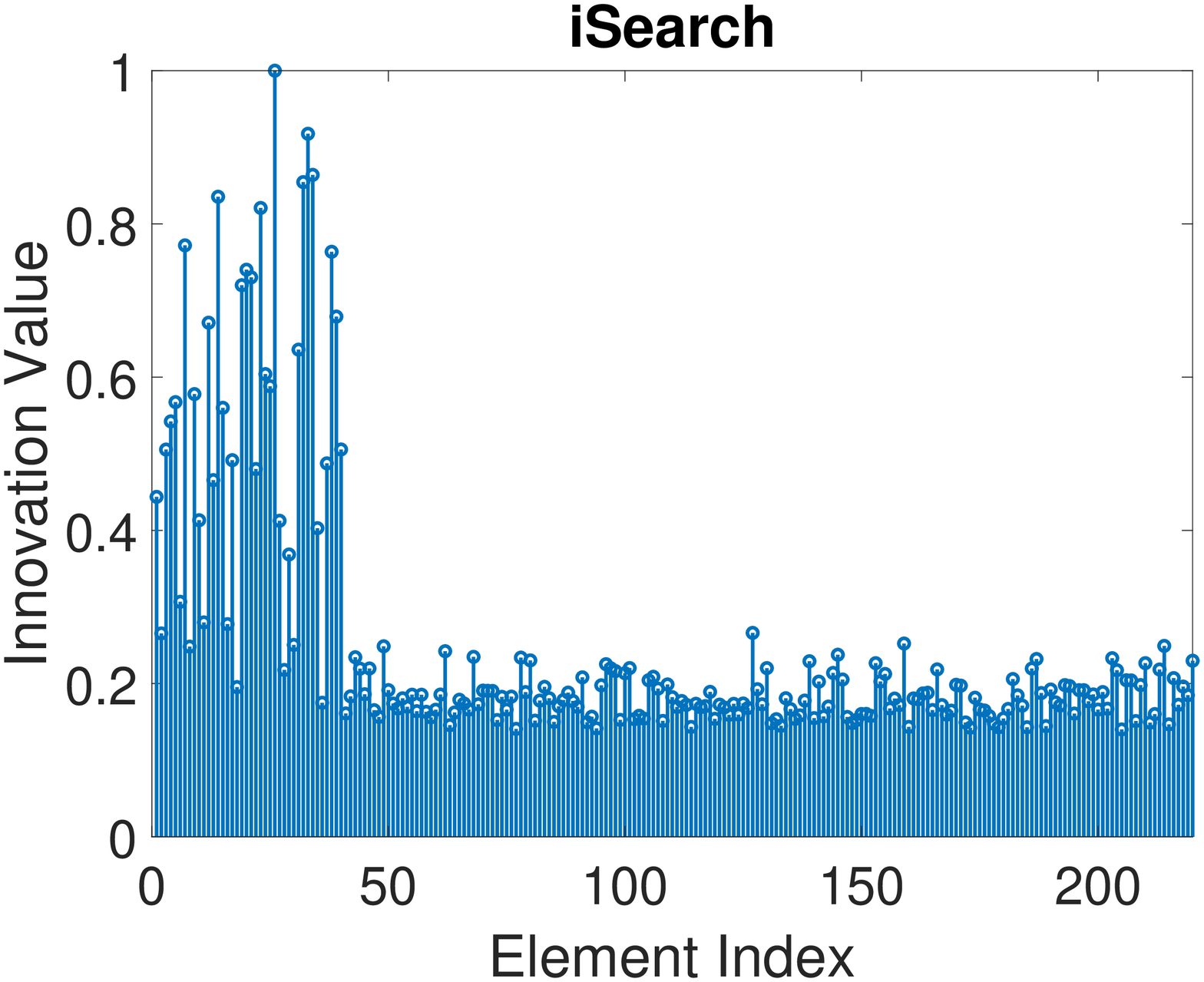}\hspace{-0.1in}
\includegraphics[width=1.78in]{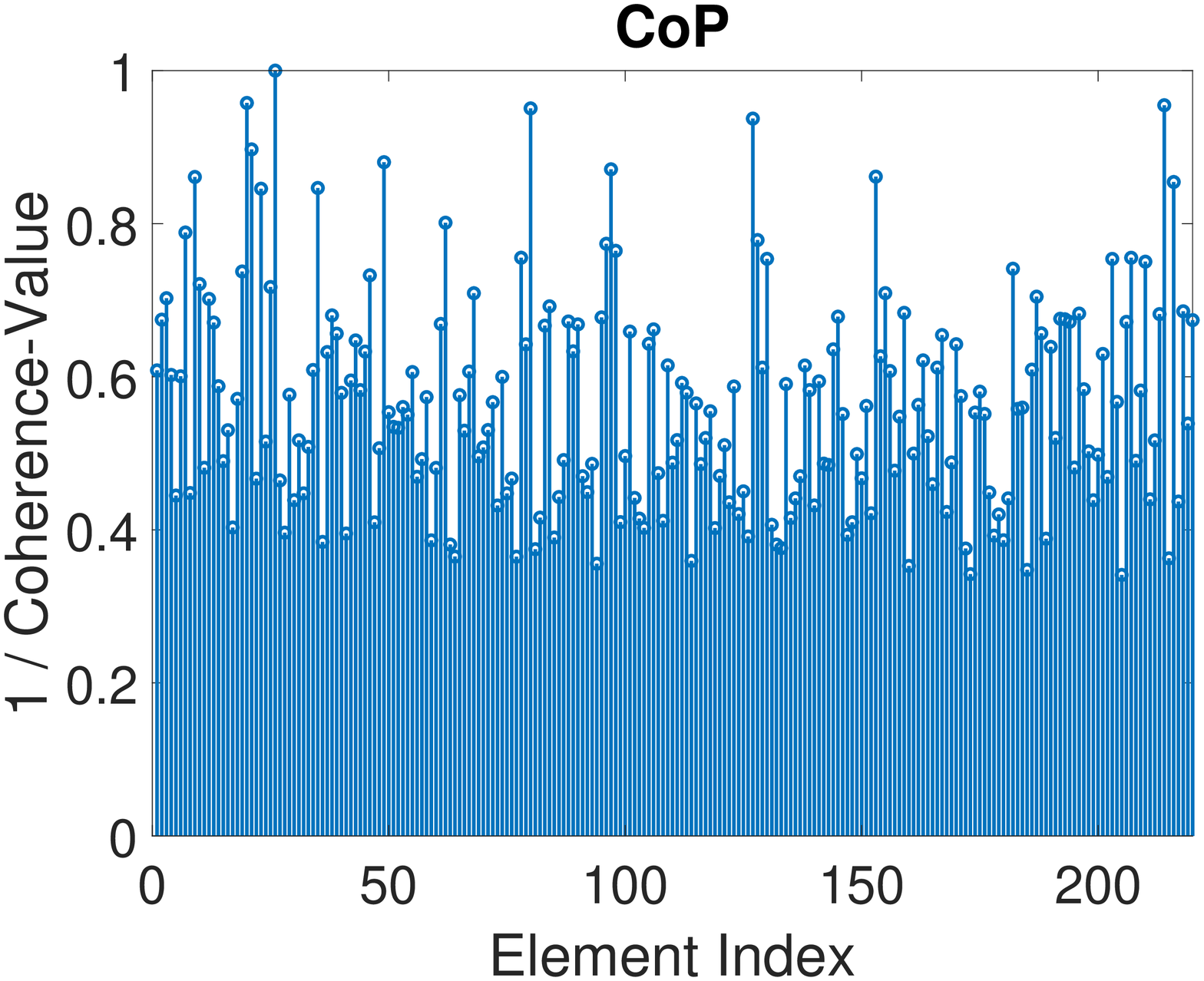}
}
\end{center}
\vspace{-0.15in}
           \caption{The plots  show the inverse of Normalized Coherence Values computed by SNCP and ANCP, the Innovation Values, and the inverse of Coherence  Values. In this experiment, the first 40 columns are the outliers. }
    \label{fig:compare_scores}
\end{figure}

In this experiment, 
we simulate a scenario in which the outliers  are close to $\calU$.
Suppose $r=8$, $n_i = 180$, and $n_o = 40$. The outliers are generated as $[\bU \:\: \bH]\: \bG$ where $\bH \in \mathbb{R}^{M_1 \times 4}$ spans a random 4-dimensional subspace and the elements of $\bG \in \mathbb{R}^{12 \times 20} $ are sampled independently from $\calN(0,1)$.  Fig.~\ref{fig:compare_scores} demonstrates  Innovation Values, Coherence Values, and Normalized Coherence Values computed by ANCP and SNCP (the first 40 columns are outliers). In this figure, we show the  inverse of Coherence Values and the inverse of Normalized Coherence Values to make them comparable to Innovation Values. One can  observe that the scores computed by iSearch, ANCP, and SNCP can be reliably used to form a basis for $\calU$ but the scores computed by CoP do not distinguish the outliers well enough.
As it was predicted by Theorem~\ref{theo:Linearly_dependant_CP},  CoP  can fail to distinguish the outliers when they are close to $\calU$. The main reason is that CoP measures the similarity between the data points via a simple inner-product while iSearch and ANCP  utilize the directions of innovation to measure the similarity between a data point and the rest of  data.   The functionality of SNCP is similar to that of ANCP while it uses a symmetric measure of similarity and the plots show that it distinguishes the outliers in a more clear way.

\subsection{Noisy Data}

In this section, we examine the robustness of the robust PCA methods against noise. Suppose $\bB$ follows Assumption~\ref{asm:out}, $r=5$, $r_o=10$, $M_1 = 200$, $n_i=100$, and $n_o=100$
where $\calU_o$ is a random 10-dimensional subspace. We consider two models for the distribution of the inliers. The first model is random distribution on $\calU \cap \mathbb{S}^{N-1}$ as described in Assumption~\ref{assum_DistUni}. In the second model, it is assumed that the inliers form a cluster in $\calU$. The following assumption describes the second model. 

 \begin{assumption}
Each column of matrix $\bA$ is formed as $\ba_i = \frac{ \bU \bs_i}{\|  \bU \bs_i \|_2} $ where $\bs_i =  \bw + \gamma \bz_i$, $\bw \in \mathbb{R}^{r }$ is a unit $\ell_2$-norm vector, and $\{\bz_i \}_{i=1}^{n_i}$ are sampled randomly  from $\mathbb{S}^{r - 1}$.
\label{asm:inliers_clus}
\end{assumption}

\begin{figure}[h!] 
\begin{center}
\mbox{
\includegraphics[width=1.8in]{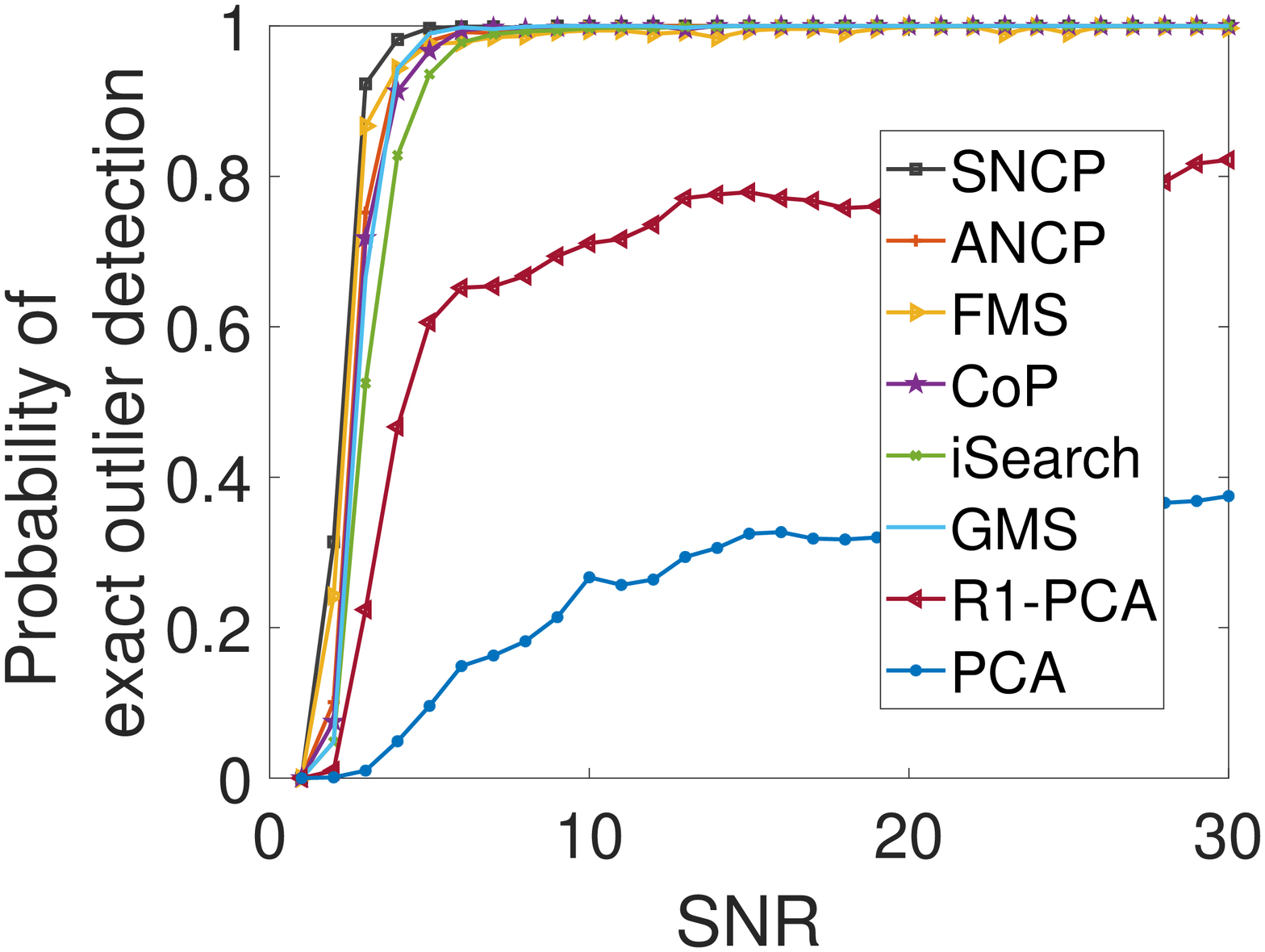}\hspace{-0.1in}
\includegraphics[width=1.8in]{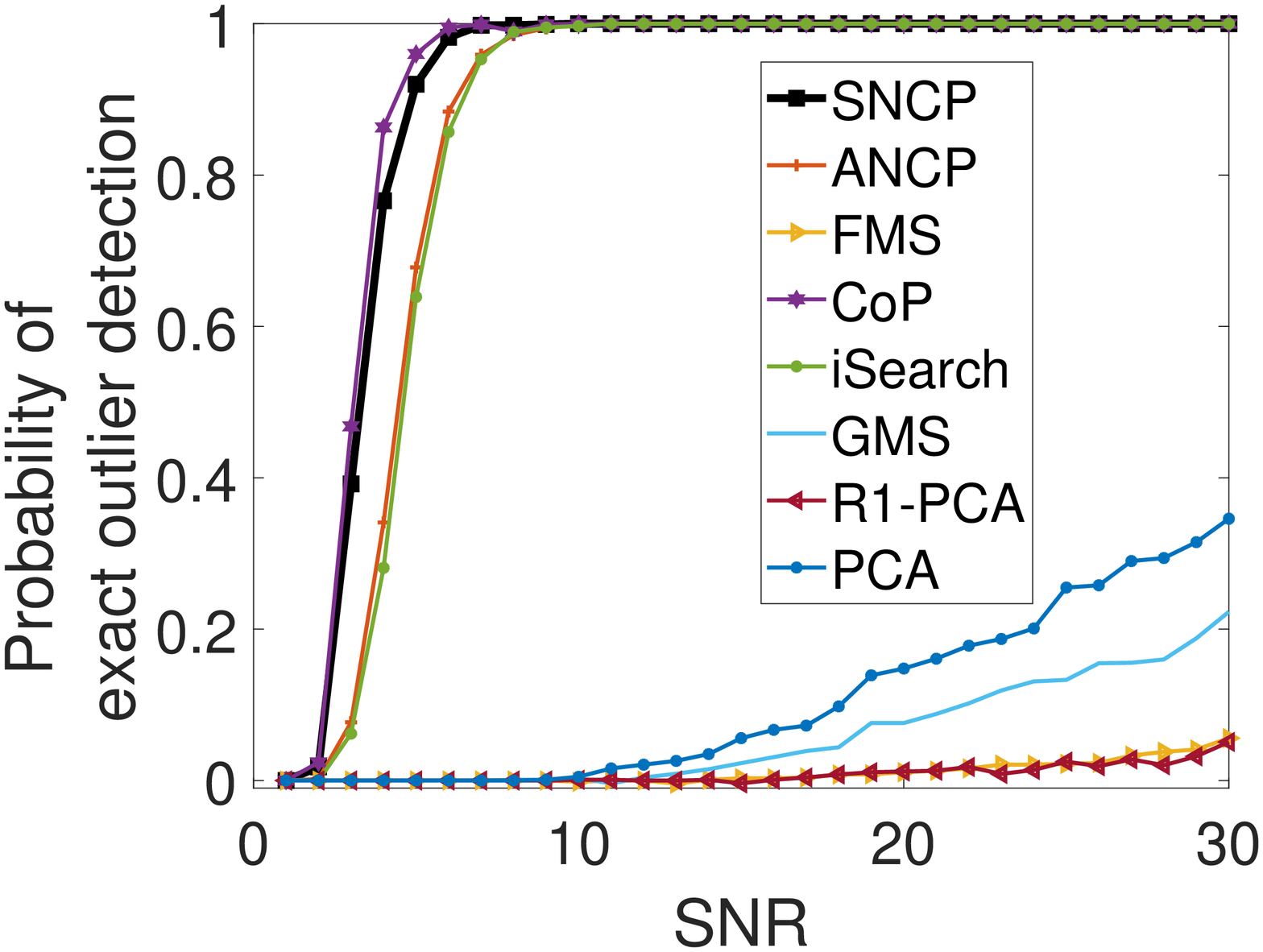}}
\end{center}
\vspace{-0.15in}
           \caption{The outliers are linearly dependant and they lie in a 10-dimensional subspace. In the left plot, the inliers are randomly distributed in $\calU$ (Assumption~\ref{assum_DistUni}) and in the right plot, the inliers form a cluster (Assumption~\ref{asm:inliers_clus} with $ \gamma = 0.2)$. }

\label{fig:lin_dep_out}
\end{figure}

\noindent
Since the data is noisy, exact subspace recovery is not feasible. Instead, we examine the probability that an algorithm distinguishes  all the outliers correctly. Define vector $\mathbf{f} \in \mathbb{R}^{M_2 }$ such that $\mathbf{f}(k) = \| (\bI - \hat{\bU} \hat{\bU}^T) \bd_k \|_2$ where $\hat{\bU}$ is the identified subspace. A trial is considered successful if
\begin{eqnarray}
\label{eq:out_conddi}
\max \bigg( \{\mathbf{f}(k) \: \: : \: \: k > n_o \} \bigg) < \min \bigg( \{\mathbf{f}(k) \: \: : \: \: k \le n_o \} \bigg),
\end{eqnarray} 
which means that the norm of projection of all the inliers on $\hat{\calU}^{\perp}$ should be smaller than the corresponding values for the outliers.  Define  $\text{SNR} = \frac{\| \bA \|_F^2}{\| \bE \|_F^2 }$ where $\bE$ is the noise component which is added to the inliers. Fig.~\ref{fig:lin_dep_out} shows the probability that (\ref{eq:out_conddi}) is valid versus SNR (the number of evaluation runs was 200). In the left plot, the distribution of inliers follows Assumption~\ref{assum_DistUni} and in the right plot it follows Assumption~\ref{asm:inliers_clus} with $\gamma = 0.2$. 
 One can observe that SNCP outperforms most of the existing methods on both cases and the performance of iSearch and ANCP are close. 
 In addition, by comparing the two plots, it can be observed that the performance of some of the robust PCA methods is sensitive to the distribution of the inliers. For instance, FMS outperforms most of the other methods when the inliers are randomly distributed but its performance degrades significantly when the inliers form a cluster in $\calU$. 

\subsection{ Identifying the Permuted Columns  }

The problem of  regression with  unknown permutation
is similar to the conventional  regression but the correspondence between input variables and labels is missing or erroneous. Suppose $\bX \in \mathbb{R}^{d \times n}$ is the measurement matrix where $n$ is the number of measurements. Define $\bY \in \mathbb{R}^{m \times n}$ as the observation matrix which can be written as $\bY = \Theta \bX$ where $\Theta \in \mathbb{R}^{m \times d}$ is the unknown matrix which is estimated by the regression algorithm. 
In the regression problem with  unknown permutation, the observation matrix $\bY$ is affected by an unknown permutation matrix $\Pi$, i.e., matrix $\bY$ can be written as $\bY = \Theta \bX \Pi$ where $\Pi \in \mathbb{R}^{n \times n}$. In this problem, it is assumed that $\Pi$ does not displace all the columns of $\Theta \bX$ and only an unknown fraction of the columns are displaced. The authors of~\cite{slawski2019sparse} showed that this special regression problem can be translated into a robust PCA problem. Define matrix $\bZ \in \mathbb{R}^{(d+m) \times n}$ as $\bZ = ([\bX^T \:\: \bY^T])^T$, i.e., each column of $\bZ$ is equal to the concatenation of the corresponding columns of $\bX$ and $\bY$. 
Suppose $n > d$ and assume that the rank of $\bX$ is equal to $d$. 
If $\Pi$ is equal to the identity matrix, the rank of $\bZ$ is equal to $d$. In contrast, when the columns of $\bY$ are displaced, the corresponding columns of $\bZ$ do not lie in the $d$-dimensional subspace which the other columns of $\bZ$ lie in. Therefore,  the columns of $\bZ$ which are corresponding to the displaced columns can be considered as outliers and a robust PCA method can be utilized to locate them. Once they are located and removed, the regression problem can be solved using the remaining measurements. 

\begin{figure}[h!] 
\begin{center}
\mbox{
\includegraphics[width=2.1in]{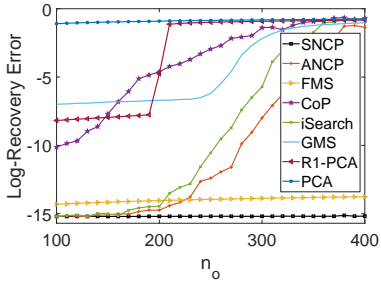}
    }
\end{center}
\vspace{-0.15in}
           \caption{This plot shows subspace recovery error versus the number of displaced measurements. The number of measurements which are not displaced are equal to $n_i=200$ and the total number of measurements are equal to $n_i + n_o$.}

\label{fig:lin_dep_out2}
\end{figure}

In this experiment, the elements of $\bX$ and $\Theta$ are sampled from $\calN (0,1)$, $d = 10$, and $m = 10$. Define $n_i$ as the number of columns of $\bY$ which are not affected by the permutation matrix and define $n_o$ as the number of displaced columns. The robust PCA methods are applied to $\bZ$ to find a basis for the $d$-dimensional subspace which is spanned by the columns of $\bZ$ corresponding to the inliers. If this subspace is estimated accurately, all the displaced columns can be exactly located~\cite{slawski2019sparse}.  Define Log-Recovery Error   as
           $
\log_{10} \left( \frac{\| (\bI - \bU \bU^T) \hat{\bU} \|_F}{ \|\bU \|_F } \right)
$, where $\hat{\bU}$ is an orthonormal basis for the recovered subspace. 
Fig.~\ref{fig:lin_dep_out2} shows Log-Recovery Error versus $n_o$ where $n_i$ is fixed equal to 200 (the number of evaluation runs was 400). 
This is a challenging subspace recovery task because the outliers can be close to the span of inliers and this is the main reason that CoP did not perform well. One can observe that SNCP and FMS yielded the best performance.  

\subsection{Unstructured Outliers}
\begin{figure}[h!]
\begin{center}
\mbox{
\includegraphics[width=0.5\textwidth]{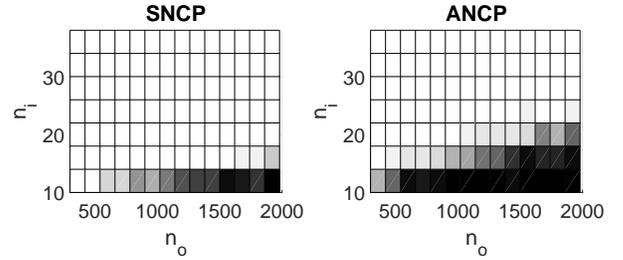}
}
\end{center}
\vspace{-0.15in}
           \caption{ The  phase transitions in presence of the unstructured outliers versus $n_i$ and $n_o$. White indicates correct
subspace recovery and black designates incorrect recovery. In this experiment, the data follows Assumption~\ref{assum_DistUni} with $M_1=50$ and $r=4$. }
\label{fig:phase}
\end{figure}

Theorem~\ref{theo:random} predicted that when the outliers are randomly distributed, the number of outliers can be much larger than the number of inliers provided that $n_i/r$ is sufficiently large. Suppose the data follows Assumption~\ref{assum_DistUni} with $M_1=50$ and $r=4$.
Define $\hat{\bU}$ as an orthonormal basis for the recovered subspace. A trial is considered successful if
$
\frac{\| (\bI - \bU \bU^T) \hat{\bU} \|_F}{ \|\bU \|_F } < 10^{-3} \: .
$
Fig.~\ref{fig:phase} shows the phase transitions 
in which white means correct subspace recover and black  designates  incorrect  recovery (the number of evaluation runs was 20). The phase transitions indicate that  when $n_i/r$ is larger than 5, the  algorithms can successfully recover $\calU$ even if $n_o = 2000$. In addition, SNCP shows more robustness against the outliers when $n_i$ is small.

\subsection{Structured Outlier Detection in Real Data}
The authors of~\cite{gitlin2018improving,rahmani2017coherence} proposed to use robust PCA to improve the accuracy of the clustering algorithms. The robust PCA method is applied to each identified cluster to find the miss-classified data points as outliers. We refer the reader to~\cite{gitlin2018improving,rahmani2017coherence}  for further details. Similar to corresponding experiment in~\cite{rahmani2017coherence}, we use Hopkins155 dataset which makes the outlier detection problem challenging. In this dataset, the data points are linearly dependent and the clusters are  close to each other. Therefore, the outliers are  structured and they are close to the inliers. The clustering error of the clustering algorithm is $30 \%$ and we compute the final clustering error after applying the robust PCA methods and updating the clusters. Table~\ref{tab:accuracy} shows the clustering error after applying different robust PCA methods. One can observe that iSearch, ANCP, SNCP, and CoP yielded better performance and the main reason is that they leverage the clustering structure of the inliers and they are robust against structured outliers. 

\begin{table}[h!]
\centering
\caption{Clustering error after using the robust PCA methods to detect the misclassified data points. }
\begin{tabular}{| c | c | c  | c |c|c|c|c|}
\hline
 CoP & FMS  & R1-PCA & GMS &     iSearch  & PCA  & ANCP & SNCP \\
 \hline
 6.93 & 28.5  & 22.56 & 17.25 &     3.72  & 12.01  & 6.64 & 3.65 \\
 \hline
\end{tabular}
\label{tab:accuracy}
\end{table}

\begin{figure*}[t]
\begin{center}
\mbox{
\includegraphics[width=1.22in]{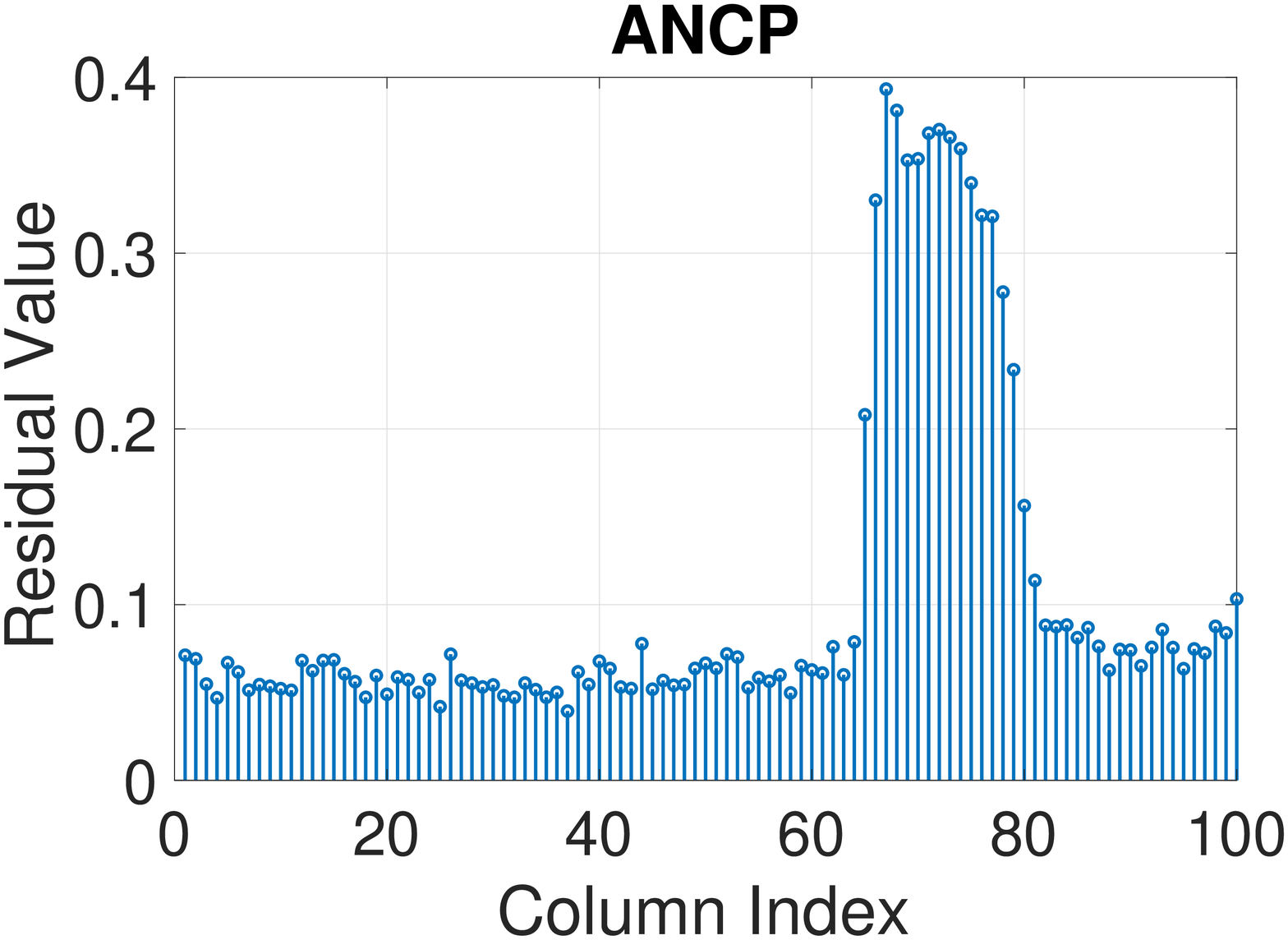}\hspace{-0.12in}
\includegraphics[width=1.22in]{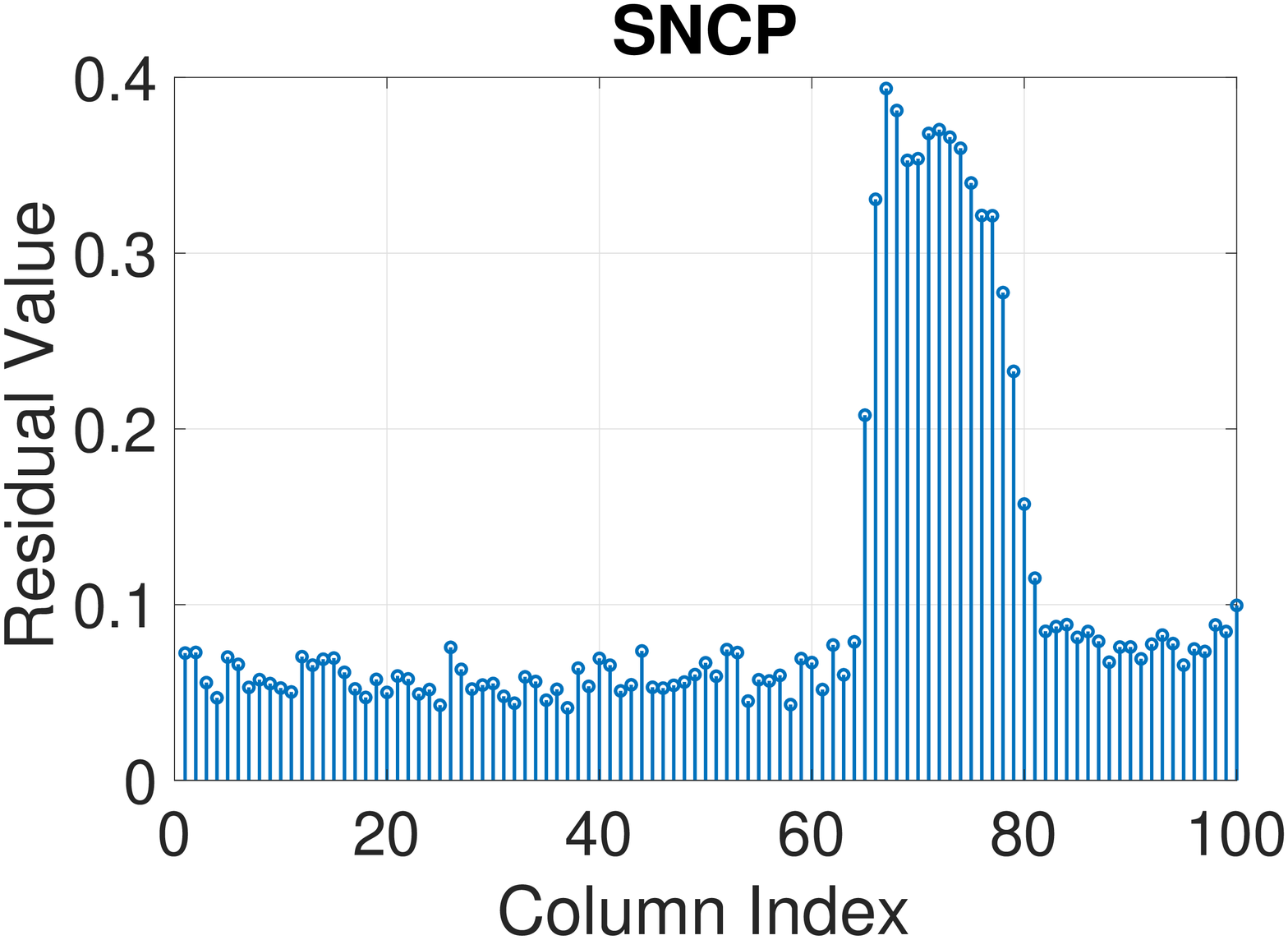}\hspace{-0.12in}
\includegraphics[width=1.22in]{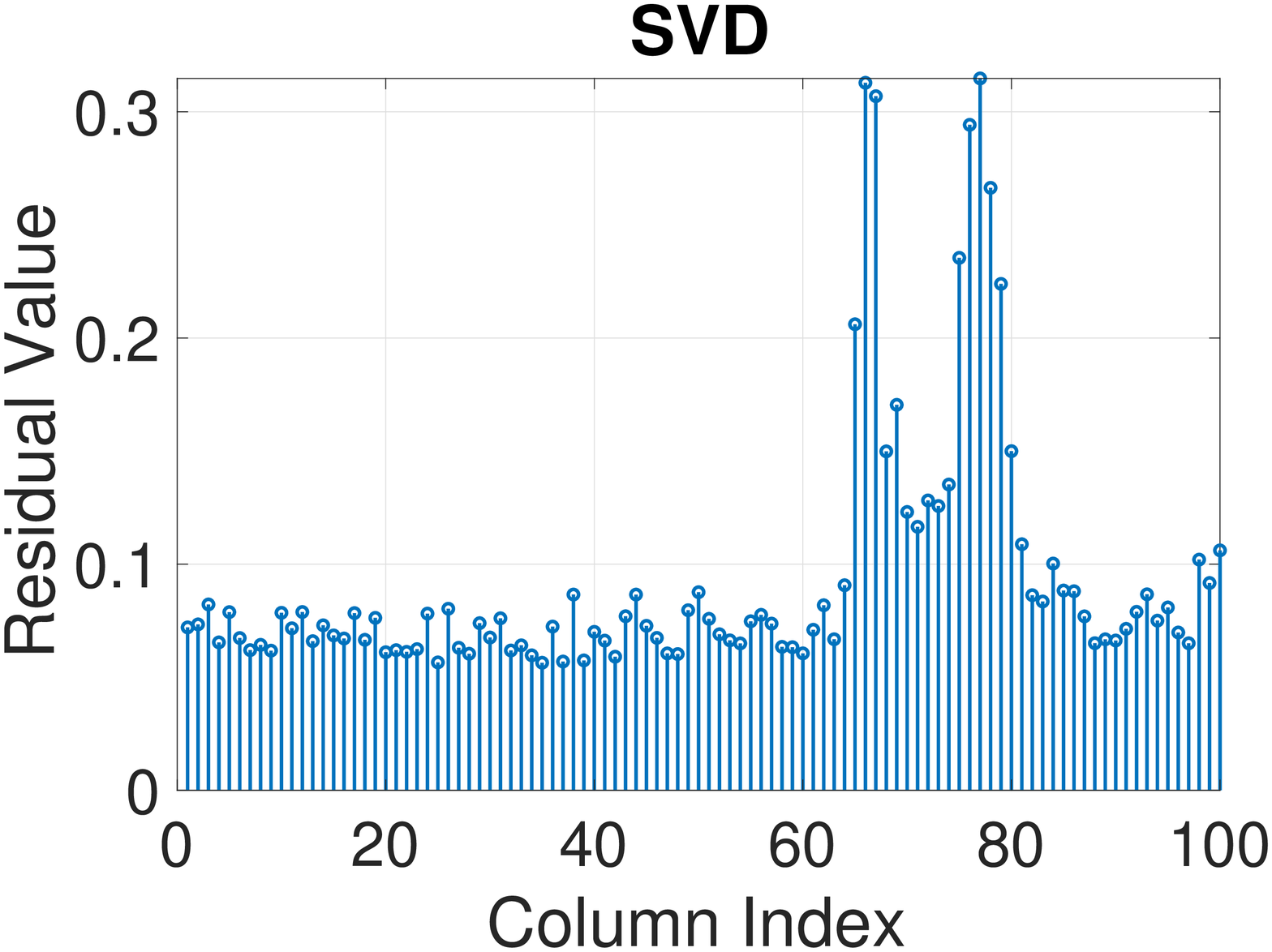}\hspace{-0.12in}
\includegraphics[width=1.22in]{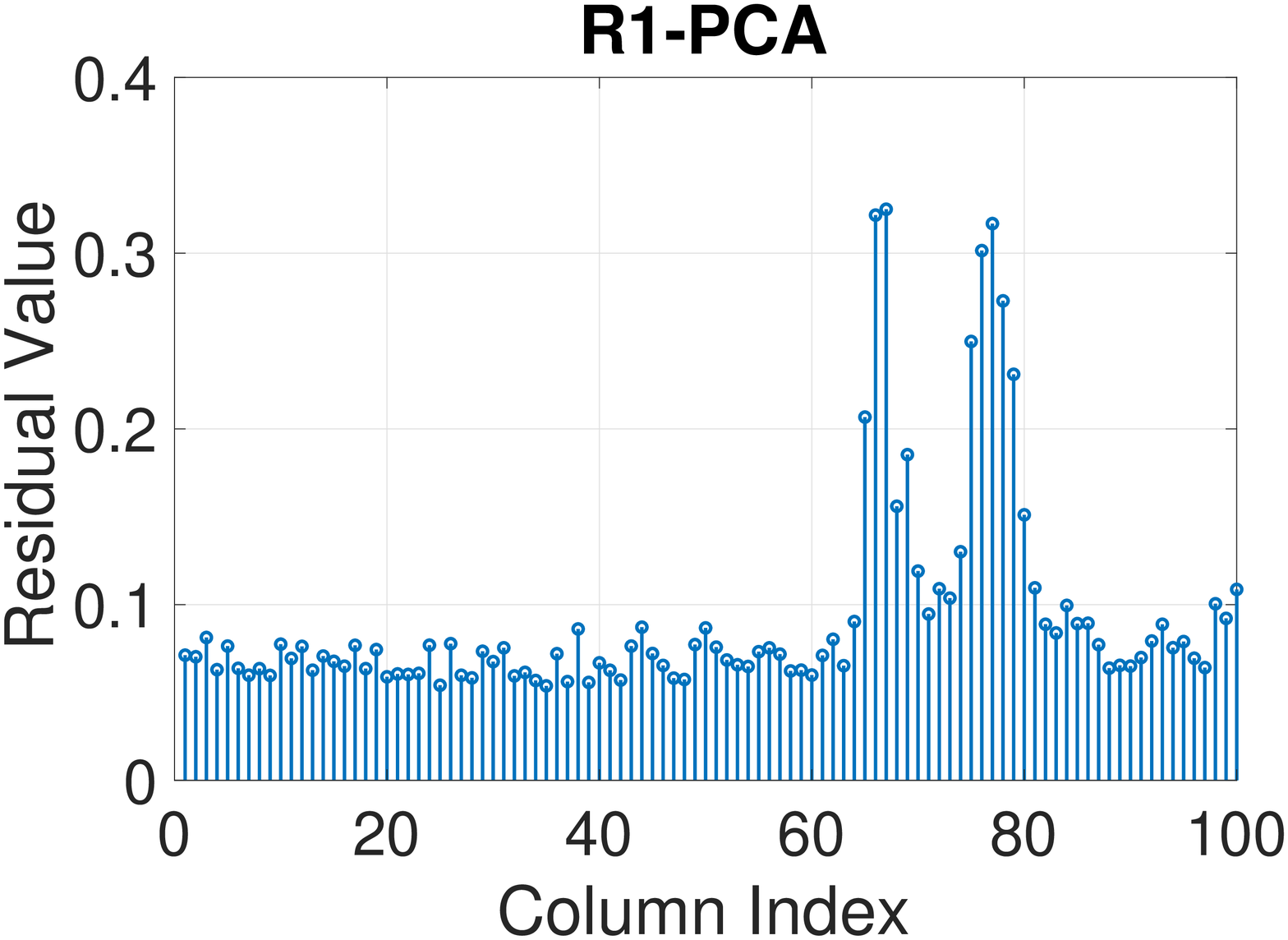}\hspace{-0.12in}
\includegraphics[width=1.22in]{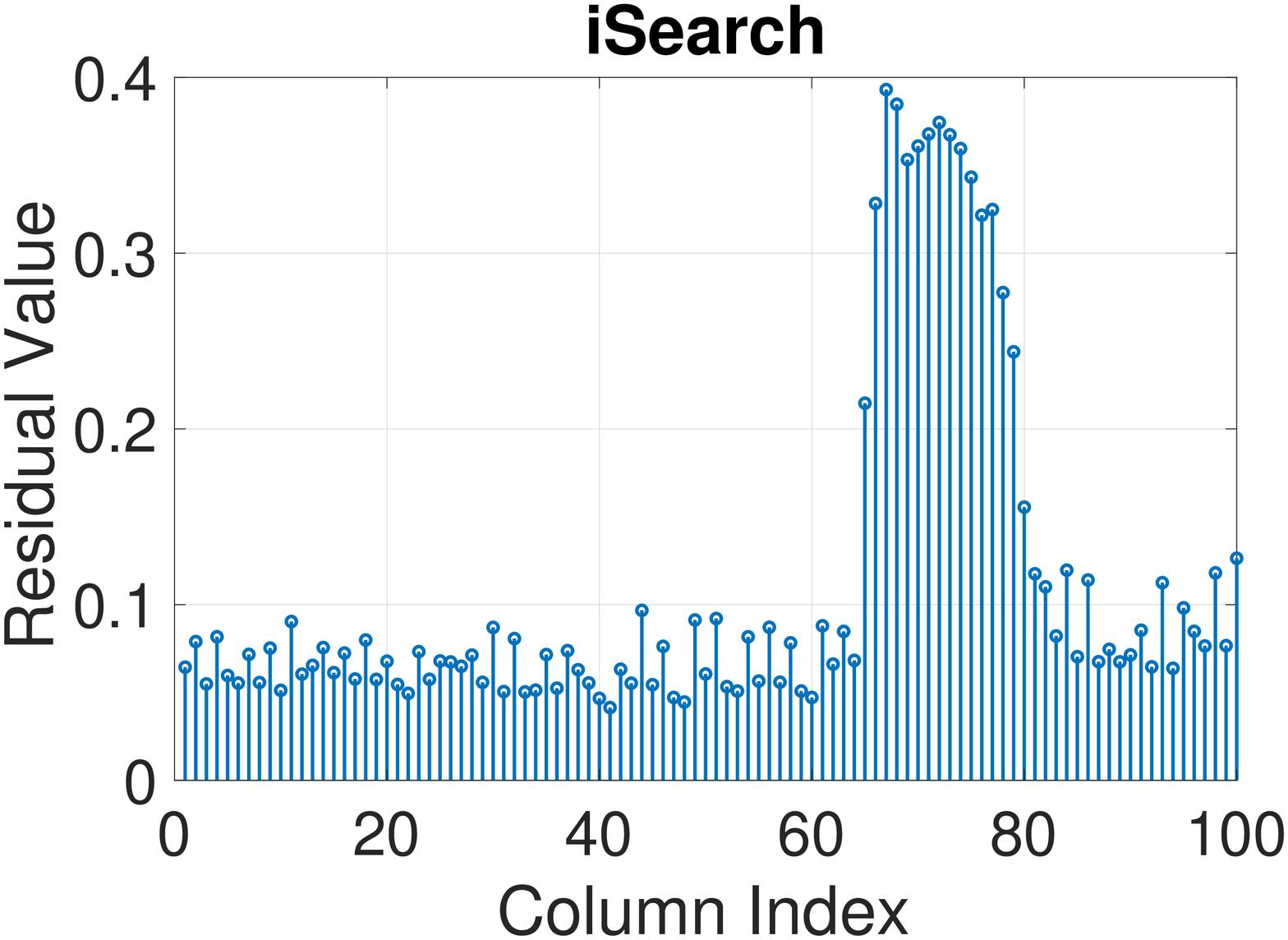}\hspace{-0.12in}
\includegraphics[width=1.22in]{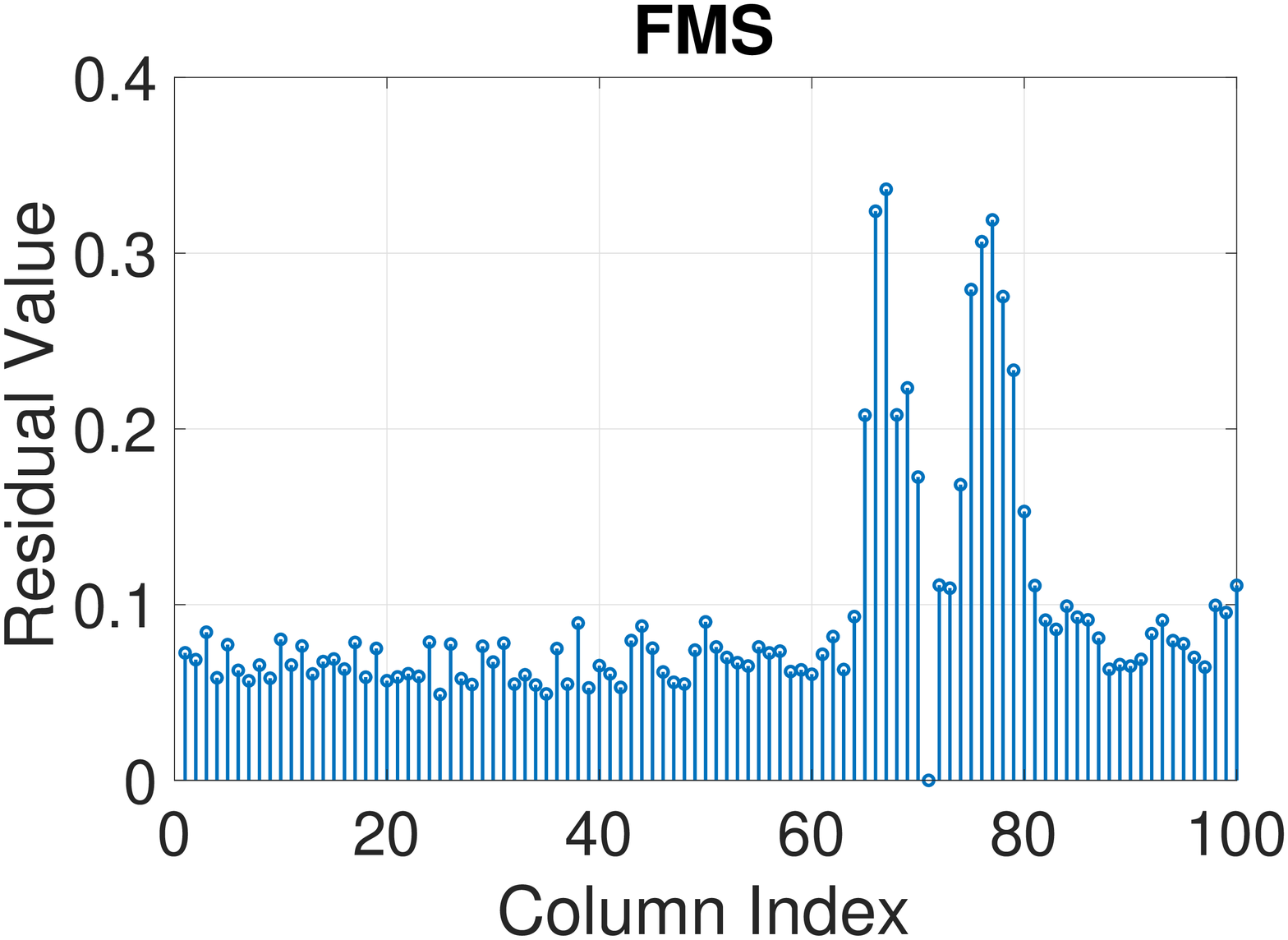}
}
\end{center}
\vspace{-0.15in}
           \caption{\textcolor{black}{This figure shows the residual} values computed by different methods. Each element represents a frame of the video file and frames 65 to 80 are the outlying frames.  }
    \label{fig:activity}
\end{figure*}

\subsection{Event Detection in Video}
In this experiment, we utilize the robust PCA methods to identify an activity in a video files, i.e., the outlier detection methods identify the frames which contain the activity as the outlying frames/data-points. 
We use the Waving Tree video file~\cite{li2004statistical} where in this video
 a tree is smoothly waving and in
the middle of the video, a person crosses the frame. The frames which only contain the background (the tree and the environment) are inliers and the few frames corresponding to the event (the presence of the person) are the outliers.
The tree is smoothly waving and we use $r=3$ as the rank of inliers for all the methods.  We use 100 frames where frames 65 to 80 are the outlying frames. In this interval (65 to 80), the person enters the frame from left, stay in the middle, and leaves from right. In this experiment, we vectorize each frame and form data matrix $\bD$ by the vectorized frames. In addition, we reduce the dimensionality of $\bD$ by projecting each column of $\bD$ into the span of the first 50 left singular vectors. Thus, $\bD \in \mathbb{R}^{50 \times 100}$.
Define $\hat{\bU}$ as the estimated subspace and define the residual value corresponding to data point $\bd_i$ as ${\| \bd_i - \hat{\bU}\hat{\bU}^T \bd_i\|_2}$. The outliers are detected as the data points with the larger residual values.
Fig.~\ref{fig:activity} shows the residual values computed by different methods. An important observation is that FMS, PCA (SVD), and R1-PCA clearly distinguished the first and the last outlying frames but they hardly distinguish the middle outliers (frames 69 to 74). The main reason is that in these frames the person does not move which means that these outlying frames are very similar to each other. Fig.~\ref{fig:activity} shows that the ANCP, SNCP, and iSearch successfully distinguish all the outlying frames since they are robust to structured and linearly dependant outliers. 

\subsection{Running Time}
In this section, we study the running time of the robust PCA methods. For ANCP, SNCP, CoP, and iSearch, we used 50 data points to build the basis matrix (matrix $\bY$). Table
\ref{tab:runnig_M2} shows the running times versus $M_2$ while $M_1 = 200$. Table~\ref{tab:runnig_M1} shows the  running times versus $M_1$ while $M_2 = 1500$. In all the runs, $r = 5$ and $n_i = 200$. One can observe that CoP, SNCP, and ANCP are notably fast since they are single step algorithms. The running time of CoP and SNCP is longer than ANCP when $M_2$ is large because their computation complexity scale with $M_2^2$. GMS is also  a fast algorithm when $M_1$ is small but its running time can be long when $M_1$ is large because its computation complexity scale with $M_1^3$.

\begin{table}[h!]
\centering
\caption{Running time of the algorithms versus $M_2$ ($M_1 = 200$).}
\begin{tabular}{| c | c | c  | c |c|c|c|}
\hline
$M_2 $& SNCP & ANCP & iSearch & CoP & FMS & GMS \\
 \hline
500 & 0.0228 & 0.0120 & 0.2660 & 0.016 & 0.1130 &  0.0872\\
 \hline
 1000 & 0.0427 & 0.0160 & 0.8983 & 0.0325 & 0.2440 & 0.1265\\
 \hline
  5000 & 0.2622 & 0.0428 & 19.4080 & 0.3930 & 0.6635  & 0.2926\\
 \hline
\end{tabular}
\label{tab:runnig_M2}
\end{table}

\begin{table}[h!]
\centering
\caption{Running time of the algorithms versus $M_1$ ($M_2 = 1500$).}
\begin{tabular}{| c | c | c  | c |c|c|c|}
\hline
$M_1 $& SNCP & ANCP & iSearch & CoP & FMS & GMS \\
 \hline
200 & 0.0614 & 0.0187 & 1.7279 & 0.0576 & 0.2978 &  0.1458 \\
 \hline
 500 & 0.1456  & 0.0727  & 2.1261  & 0.0710  & 0.9574 & 0.6399\\
 \hline
  1000 & 0.3145 & 0.2527  & 2.7695  & 0.0900  & 2.7731  & 2.5590 \\
 \hline
\end{tabular}
\label{tab:runnig_M1}
\end{table}

\section{Conclusion}
It was shown that  Innovation Value under the quadratic cost function is equivalent to  Leverage Score. Two closed-form robust PCA methods were presented where the first one was based on Leverage Score and the second one was inspired by the connection between Leverage Score and Innovation Value. Several theoretical performance guarantees for the robust PCA method under different models for the distribution of the outliers and the distribution of the inliers were presented. In addition, it was shown with both theoretical and numerical investigations that the algorithms are robust to the strong presence of noise. Although the presented methods are fast  closed-form algorithms, it was shown that they often outperform most of the existing methods.

\vspace{0.4in}

 \bibliography{example_paper}
 \bibliographystyle{plain}

\newpage
\section{Proofs}
In this section, the proofs for the presented theoretical results are provided. In most of the proofs, we frequently use the following useful lemmas.

\begin{lemma}
\cite{rahmani2017coherence}
Suppose $\bg_1, ... ,\bg_{n}$   are i.i.d. random vectors distributed uniformly on the unit sphere $\mathbb{S}^{N - 1}$ in $\mathbb{R}^{N  }$. If $N > 2$, then
\begin{eqnarray*}
\begin{aligned}
& \underset{\|\bu\| = 1}{\sup} \:\: \sum_{i = 1}^{n} (\bu^T \bg_i)^2 \leq \frac{n}{N} + \max \left( \frac{4}{3} \log \frac{2N}{\delta} , \sqrt{4 \frac{n}{N} \log \frac{2 N}{\delta}} \right)  \\
& \underset{\|\bu\| = 1}{\inf} \:\: \sum_{i = 1}^{n} (\bu^T \bg_i)^2 \ge \frac{n}{N} - \max \left( \frac{4}{3} \log \frac{2N}{\delta} , \sqrt{4 \frac{n}{N} \log \frac{2 N}{\delta}} \right) 
\end{aligned}
\end{eqnarray*}
with probability at least $1 - \delta$.
\label{lm:mylemma}
\end{lemma}

\begin{lemma}
\cite{lamport49} Suppose orthonormal matrix $\bF \in \mathbb{R}^{N \times r}$ spans a random $r$-dimensional  subspace. For a given vector $\bc \in \mathbb{R}^{N \times 1}$
\begin{eqnarray*}
\mathbb{P} \left[ \| \bc^T \bF \|_2 > \sqrt{\frac{c_1 \Bar{r}}{N}}  \right] \leq 1 - c_2 N^{-3} \log N  \:,
\end{eqnarray*}
where $c_1$ and $c_2$ are constant real numbers and $\Bar{r} = \max (r , \log N)$.
\label{lm:random_project}
\end{lemma}

\begin{lemma}
\cite{rahmani2017coherence}
Suppose $\bg_1, ... ,\bg_{n}$   are i.i.d. random vectors distributed uniformly on the unit sphere $\mathbb{S}^{N - 1}$ in $\mathbb{R}^{N  }$. If $N > 2$, then
\begin{eqnarray}\notag
\underset{\|\bu\| = 1}{\sup} \:\: \sum_{i = 1}^{n} | \bu^T \bg_i | <  \frac{n}{\sqrt{N}} + 2\sqrt{n} + \sqrt{\frac{2 n \log \frac{1}{\delta}}{N -1 }}
\end{eqnarray}
with probability at least $1 - \delta$.
\label{lm:perm_positive}
\end{lemma}

\subsection{Proof of Lemma~\ref{lem:equi}}
The optimal point of (\ref{eq:el_2_inno}) is equivalent to the optimal point of \begin{eqnarray}\notag
\underset{ \bc}{\min} \: \:  \bc^T \bD\bD^T \bc \quad \text{subject to} \quad \bc^T \bd_i = 1 
\end{eqnarray}
whose Lagrangian function  is as follows
\begin{eqnarray}\notag
\bc^T \bD\bD^T \bc + \gamma( \bc^T \bd_i - 1) \:,
\end{eqnarray}
where $\gamma$ is the Lagrangian multiplier. 
Assuming that $\bD$ is a full rank matrix, the Lagrangian function can be used to find the optimal point  as 
\begin{eqnarray}
\begin{aligned}
    \bc_i^{*} = \frac{(\bD \bD^T)^{-1} \: \bd_i}{\bd_i^T \: (\bD \bD^T)^{-1} \: \bd_i} \:.
\end{aligned}
\label{eq:cistar}
\end{eqnarray}
Accordingly,
\begin{eqnarray}\notag
\begin{aligned}
  &  \| \bD^T \bc_i^{*} \|_2^2  = \frac{\bd_i^T (\bD \bD^T)^{-1} \bD \bD^T (\bD \bD^T)^{-1} \: \bd_i}{(\bd_i^T \: (\bD \bD^T)^{-1} \: \bd_i)^2} = \\
   &\quad \quad\frac{1}{\bd_i^T \: (\bD \bD^T)^{-1} \: \bd_i} \:.
\end{aligned}
\end{eqnarray}
Each data point $\bd_i$ can be written as $\bU^{'} \Sigma \bv_i$. Thus,
\begin{eqnarray}\notag
\begin{aligned}
 \| \bD^T \bc_i^{*} \|_2^2 = \frac{1}{\bd_i^T \: (\bD \bD^T)^{-1} \: \bd_i} = \frac{1}{\bv_i^T \bv_i} \:.
\end{aligned}
\end{eqnarray}

\subsection{Proof of Lemma~\ref{lm:secondlm}}
According to (\ref{eq:cistar}), 
\begin{eqnarray*}
\begin{aligned}
 \| \bD^T \bc_i^{*} \|_2^2 & = \left\| \bD^T \frac{(\bD \bD^T)^{-1} \: \bd_i}{\bd_i^T \: (\bD \bD^T)^{-1} \: \bd_i} \right\|_2^2 \\
 & =\sum_{i=j}^{M_2} \left( \frac{\bd_i^T (\bD \bD^T)^{-1} \bd_j}{\bd_i^T \: (\bD \bD^T)^{-1} \: \bd_i} \right)^2\:.
\end{aligned}
\end{eqnarray*}
Each data point $\bd_i$ can be written as $\bU^{'} \Sigma \bv_i$. Therefore, 
\begin{eqnarray*}
\begin{aligned}
 \| \bD^T \bc_i^{*} \|_2^2 = \sum_{i=j}^{M_2} \left( \frac{\bv_i^T \bv_j}{\bv_i^T \bv_i} \right)^2\:.
\end{aligned}
\end{eqnarray*}

\subsection{Proof of Theorem~\ref{theo:randomrandom}}
Define $\bv$ as the columns of $\bV \in \mathbb{R}^{r_d \times M_2}$ (the matrix of right singular vectors) which is corresponding to data point $\bd$.
The defined Normalized Coherence Value can be written as
\begin{eqnarray}\notag
\begin{aligned}
 {\bc^{*}}^T \bH {\bc^{*}}
\end{aligned}
\end{eqnarray}
where $\bH = \bD \bD^T$ and $\bc^{*}$ is the optimal point of
\begin{eqnarray}
\underset{ \bc}{\min} \: \:  \bc^T \bH \bc \quad \text{subject to} \quad \bc^T \bd = 1 \:.
\label{eq:first11}
\end{eqnarray}

In order to guarantee exact recovery of $\calU$, it is enough to show that (\ref{cond:main_cond}) holds. Accordingly,  we establish  a lower bound for the Normalized Coherence Values corresponding to the inliers and an upper-bound for the Normalized Coherence Values corresponding to the outliers and we derive the sufficient conditions to guarantee that the lower-bound is larger than the upper-bound. 


Suppose $\bd$ is an inlier. Then, the linear constraint of (\ref{eq:first11}) ensures that $\| \bc^{*} \|_2 \ge 1$ and $\| \bU^T \bc^{*} \|_2 \ge 1$. Therefore, 
\begin{eqnarray} 
\begin{aligned}
  {\bc^{*}}^T \bH \bc^{*}  & = \|\bA^T \bc^{*} \|_2^2 + \|\bB^T \bc^{*} \|_2^2 \\ 
 & \ge \| \bU^T \bc^{*}\|_2^2 \underset{\delta \in \calU  \atop \| \delta \|_2 = 1}{\inf} \| \delta^T \bA \|_2^2 + \|\bc^{*}\|_2^2 \underset{ \delta \|_2 = 1}{\inf} \| \delta^2 \bB\|_2^2 \\
& \ge  \underset{\delta \in \calU  \atop \| \delta \|_2 = 1}{\inf} \| \delta^T \bA \|_2^2 +  \underset{ \delta \|_2 = 1}{\inf} \| \delta^2 \bB\|_2^2 \:.
\end{aligned}
\label{eq:left_1jj}
\end{eqnarray}
We can use Lemma~\ref{lm:mylemma} to establish a lower bound for the Normalized Coherence  Values corresponding to the inliers as follows 
\begin{eqnarray} 
\begin{aligned}
 & \mathbb{P} \Bigg[ \| \bD^T \bc^{*} \|_2^2  < \frac{n_i}{r} - \max \left( \frac{4}{3} \log \frac{2r}{\delta} , \sqrt{4 \frac{n_i}{r} \log \frac{2 r}{\delta}} \right)  \\
 & \hspace{.3cm} +  \frac{n_o}{M_1} - \max \left( \frac{4}{3} \log \frac{2M_1}{\delta} , \sqrt{4 \frac{n_o}{M_1} \log \frac{2 M_1}{\delta}} \right)  \Bigg] < 2\delta \:.
\end{aligned}
\label{eq:suff_1_1}
\end{eqnarray}

Now, we need to establish an upper-bound for the Normalized Coherence Values corresponding to the outliers. Define $\bd^{\perp} = \frac{\bR \bR^T \bd}{\| \bd^T \bR \|_2^2}$ where $\bR$ is an orthonormal basis for $\calU^{\perp}$ (which was defined as the complement of $\calU$). By definition, 
$$\bd^T \bd^{\perp}  = \frac{\bd^T \bR \bR^T \bd}{\| \bd^T \bR \|_2^2} = 1\:.$$
Since $\bc^{*}$ is the optimal point and $\bd^{\perp}$ satisfies the linear constraint,
\begin{eqnarray}\notag
\| \bD^T \bc^{*}\|_2^2 \le \| \bD^T \bd^{\perp} \|_2^2 = \| \bB^T \bd^{\perp}  \|_2^2 \:.
\end{eqnarray}
In addition, according to the definition of $\bd^{\perp}$, 
\begin{eqnarray}\notag
\| \bd^{\perp} \|_2 = \frac{1}{\| \bd^T \bR\|_2}\:.
\end{eqnarray}
Thus,  we can conclude that 
\begin{eqnarray}
\begin{aligned}
&\mathbb{P} \Bigg[ \|\bD^T \bc^{*} \|_2^2 >   \frac{ n_o}{\| \bd^T \bR\|_2^2 M_1} + \\
& \frac{1}{\| \bd^T \bR\|_2^2}\max \left( \frac{4}{3} \log \frac{2M_1}{\delta} , \sqrt{\frac{n_o}{M_1} \log \frac{2 M_1}{\delta}} \right)\Bigg] < \delta \:.
\end{aligned}
\label{eq:suff_2_1}
\end{eqnarray}

Therefore, according to (\ref{eq:suff_1_1}) and (\ref{eq:suff_2_1}),  if (\ref{eq:suff_1}) is satisfied, then (\ref{cond:main_cond}) holds and $\calU$ is recovered exactly with probability at least $1 - 3\delta$.

\subsection{Proof of Theorem~\ref{theo:Linearly_dependant}}
The procedure to prove Theorem~\ref{theo:Linearly_dependant} is similar to the procedure used to prove Theorem~\ref{theo:randomrandom}, i.e., we guarantee that a lower-bound for the Normalized Coherence Values corresponding to the inliers is larger than an upper-bound for the  Normalized Coherence Values corresponding to the outliers. First we establish the lower-bound for the   Normalized Coherence Values corresponding to the inliers. 
A Normalized Coherence Value can be written as 
$$ {\bc^{*}}^T \bH {\bc^{*}}  = \|\bA^T \bc^{*} \|_2^2 + \|\bB^T \bc^{*} \|_2^2 \:.$$ 
When $\bd$ is an inlier, $\|\bc^{*} \bU \|_2 \ge 1$ which means
\begin{eqnarray} 
\begin{aligned}
  \| \bD^T \bc^{*} \|_2^2  & \ge \|\bA^T \bc^{*} \|_2^2  \ge \| \bU^T \bc^{*}\|_2^2 \underset{\delta \in \calU  \atop \| \delta \|_2 = 1}{\inf} \| \delta^T \bA \|_2^2 \\
 & \ge \underset{\delta \in \calU  \atop \| \delta \|_2 = 1}{\inf} \| \delta^T \bA \|_2^2 \:,
\end{aligned}
\label{eq:left_1}
\end{eqnarray}
where $\bc^{*}$ was defined as the optimal point of (\ref{eq:first11}).
Since the inliers are randomly distributed on $\calU \cap	\mathbb{S}^{M_1 - 1}$, according to Lemma~\ref{lm:mylemma}
\begin{eqnarray} 
\begin{aligned}
 & \mathbb{P} \Bigg[ \| \bD^T \bc^{*} \|_2^2  < \\
 & \hspace{1.3cm}\frac{n_i}{r} -   \max \left( \frac{4}{3} \log \frac{2r}{\delta} , \sqrt{4 \frac{n_i}{r} \log \frac{2 r}{\delta}} \right)  \Bigg] < \delta \:.
\end{aligned}
\label{eqqqq}
\end{eqnarray}

In order to establish the upper-bound for the  Normalized Coherence Values corresponding to the outliers, first we define $\bd^{\perp} = \frac{\bR \bR^T \bd}{\| \bd^T \bR \|_2^2}$ where $\bR$ is an orthonormal basis for $\calU^{\perp}$. Vector $\bd^{\perp}$  lies in $\calU^{\perp}$ and $\bd^{T} \bd^{\perp} = 1$. Thus, 
\begin{eqnarray}\notag
\| \bD^T \bc^{*}\|_2^2 \le \| \bD^T \bd^{\perp} \|_2^2 = \| \bB^T \bd^{\perp}  \|_2^2 \:.
\end{eqnarray}
In addition, $ \| \bU_{o}^T \bd^{\perp} \|_2 \le \| \bd^{\perp}\|_2 \| \bU_{o}^T \bU^{\perp} \|$. Therefore, we can use Lemma~\ref{lm:mylemma} to establish the following upper-bound
\begin{eqnarray}
\begin{aligned}
&\mathbb{P} \Bigg[ \|\bD^T \bc^{*} \|_2^2 >  \| \bU_{o}^T \bU^{\perp} \|   \Bigg( \frac{\psi n_o}{r_o} +\\
& \quad\quad \quad \psi \max \left( \frac{4}{3} \log \frac{2r_o}{\delta} , \sqrt{\frac{n_o}{r_o} \log \frac{2 r_o}{\delta}} \right)\Bigg) \Bigg] < \delta \:.
\end{aligned}
\label{eqq22}
\end{eqnarray}
According to (\ref{eqqqq}) and (\ref{eqq22}), if the sufficient condition of Theorem~\ref{theo:Linearly_dependant} is satisfied,  then (\ref{cond:main_cond}) holds and $\calU$ is recovered exactly with probability at least $1 - 2 \delta$.

\subsection{Proof of Theorem~\ref{theo:Linearly_dependant_CP}}
The  Coherence Value corresponding to data point $\bd$~is~equal~to
\begin{eqnarray}\notag
\begin{aligned}
\sum_{i=1}^{M_2} (\bd^T \bd_i)^2 = \| \bd^T \bD \|_2^2 = \|\bd^T \bA\|_2^2 + \| \bd^T \bB\|_2^2 \:.
\end{aligned}
\end{eqnarray}
We use similar procedure which was used to prove Theorem~\ref{theo:Linearly_dependant}. First we establish a lower-bound for Coherence Values corresponding to the inliers. Suppose $\bd$ is an inlier. Similar to (\ref{eq:left_1}) and (\ref{eqqqq}), 
\begin{eqnarray} 
\begin{aligned}
 & \mathbb{P} \Bigg[ \| \bD^T \bd \|_2^2  <  \quad \frac{n_i}{r} - \max \left( \frac{4}{3} \log \frac{2r}{\delta} , \sqrt{4 \frac{n_i}{r} \log \frac{2 r}{\delta}} \right)  \Bigg] < \delta \:.
\end{aligned}
\label{eqqqqcp}
\end{eqnarray}

Next, we need to establish an upper-bound for  Coherence Values corresponding to the outliers. Since $\bd \in \calU_o$,
\begin{eqnarray}\notag 
\begin{aligned}
\| \bd^T \bD \|_2^2 \le \underset{\delta \in \calU_o  \atop \| \delta \|_2 = 1}{\sup} \| \delta^T \bB \|_2^2 + \|\bU^T \bU_o \|^2 \underset{\delta \in \calU  \atop \| \delta \|_2 = 1}{\sup} \| \delta^T \bA \|_2^2 \:.
\end{aligned}
\end{eqnarray}
Accordingly, using Lemma~\ref{lm:mylemma}, we can conclude that when $\bd$ is an outlier
\begin{eqnarray} 
\begin{aligned}
 & \mathbb{P} \Bigg[ \| \bD^T \bd \|_2^2  > \\
 & \|\bU^T \bU_o \|^2 \left( \frac{n_i}{r} + \max \left( \frac{4}{3} \log \frac{2r}{\delta} , \sqrt{4 \frac{n_i}{r} \log \frac{2 r}{\delta}} \right) \right) + \\ & \frac{n_o}{r_o} + \max \left( \frac{4}{3} \log \frac{2r_o}{\delta} , \sqrt{4 \frac{n_o}{r_o} \log \frac{2 r_o}{\delta}} \right)
 \Bigg] < 2\delta \: .
\end{aligned}
\label{eqqqqcp2}
\end{eqnarray}
According to (\ref{eqqqqcp}) and (\ref{eqqqqcp2}), if the sufficient condition of Theorem~\ref{theo:Linearly_dependant_CP} is satisfied, the minimum of  Coherence Values Corresponding to the inliers is larger than the maximum of  Coherence Values Corresponding to the outliers and $\calU$ is recovered exactly with probability at least $1- 3\delta$.

\subsection{Proof of Theorem~\ref{theo:clustered_outliers}}
Similar to the proofs of the previous theorems, first we establish a lower-bound for the Normalized Coherence Values corresponding to the inliers and an upper-bound for the Normalized Coherence Values corresponding to the outliers. Subsequently, the final sufficient conditions are derived to guarantee that the lower-bound in higher than the upper-bound with high probability. 

Using (\ref{eq:left_1}), we can bound the Normalized Coherence Values corresponding to the inliers as stated in (\ref{eqqqq}).  

Similar to the proof of Theorem~\ref{theo:Linearly_dependant}, we define  $\bd^{\perp} = \frac{\bR \bR^T \bd}{\| \bd^T \bR \|_2^2}$ where $\bR$ is an orthonormal basis for $\calU^{\perp}$. Since $\bd^{T} \bd^{\perp}=1$,
\begin{eqnarray}\notag
 {\bc^{*}}^T \bH \bc^{*} \le {\bd^{\perp} }^T \bH \bd^{\perp}  = \| \bB^T \bd^{\perp}  \|_2^2 \:.
\end{eqnarray}
In addition, 
\begin{eqnarray}\notag
\begin{aligned}
& \|  \bB^T \bd^{\perp} \|_2^2 = \frac{1  }{{1+\eta^2}} \sum_{i=1}^{n_o} \left( \bq^T \bd^{\perp} + \eta \bv_i^T \bd^{\perp} \right)^2 \le \\
&  \frac{ 1 }{{1+\eta^2}} \Bigg( n_o  ( \bq^T \bd^{\perp} )^2 + \\
& \quad\quad\quad\quad\quad\quad\eta^2 \sum_{i=1}^{n_o} \left(  \bv_i^T \bd^{\perp} \right)^2 +  2\eta | \bq^T \bd^{\perp} | \sum_{i=1}^{n_o} \left| \bv_i^T \bd^{\perp} \right| \Bigg) \:.
\end{aligned}
\end{eqnarray}
Since $\bd^{\perp}$ lies in $\calU^{\perp}$, $\bq^T \bd^{\perp} \le \|\bd^{\perp}\|_2 \: \| \bq^T \bU^{\perp} \|_2$. Moreover, according to Lemma~\ref{lm:perm_positive},
\begin{eqnarray}
\begin{aligned}
 \sum_{i=1}^{n_o} \left| \bv_i^T \bd^{\perp} \right| \le \|\bd^{\perp} \|_2 \left( \frac{n_o}{\sqrt{M_1}} + 2\sqrt{n_o} + \sqrt{\frac{2 n_o \log \frac{1}{\delta}}{M_1 -1 }} \right)
\end{aligned}
\label{eq:new_bound}
\end{eqnarray}
with probability at least $1 - \delta$.
Accordingly, using (\ref{eq:new_bound}) and Lemma~\ref{lm:mylemma}, we can establish the following bound  
\begin{eqnarray*}
\begin{aligned}
& \mathbb{P} \Bigg[ (1 + \eta^2)\|\bB^T \bd^{\perp}\|_2^2  > \psi n_o \| \bq^T \bU^{\perp} \|_2^2 +\\
& \frac{\psi \eta^2  n_o}{M_1} + \eta^2  \psi \max \left( \frac{4}{3} \log \frac{2M_1}{\delta} , \sqrt{\frac{n_o}{M_1} \log \frac{2 M_1}{\delta}} \right) + \\
&  \hspace{0cm} \eta \sqrt{\psi} \| \bq^T \bU^{\perp} \|_2 \left( \frac{n_o}{\sqrt{M_1}} + 2\sqrt{n_o} + \sqrt{\frac{2 n_o \log \frac{1}{\delta}}{M_1 -1 }} \right) \Bigg] \le 2\delta \:. 
\end{aligned}
\end{eqnarray*}
Thus, if the sufficient condition of Theorem~\ref{theo:clustered_outliers} is satisfied, then (\ref{cond:main_cond}) holds and $\calU$ is recovered exactly with probability at $1- 4 \delta$.

\subsection{Proof of Theorem~\ref{theo:Linearly_dependant_inliers}}
The only difference between this Theorem and Theorem~\ref{theo:randomrandom} is the difference between the presumed model for the distribution of the inliers. We can not use Lemma~\ref{lm:mylemma} to establish a lower-bound for $ \underset{\delta \in \calU  \atop \| \delta \|_2 = 1}{\inf} \| \delta^T \bA \|_2^2$. According to the clustering structure of the inliers,
\begin{eqnarray}
 \begin{aligned}
& \underset{\delta \in \calU  \atop \| \delta \| = 1}{\inf} \| \bA^T \delta \|_2^2 =  \underset{\delta \in \calU  \atop \| \delta \| = 1}{\inf} \sum_{i=1}^{m} \| \delta^T \bU_k \|_2^2 \left\| \bA_k^T \frac{\bU_k \bU_k^T \delta}{\| \delta^T \bU_k \|_2} \right\|_2^2 \ge \\
& \underset{\delta \in \calU  \atop \| \delta \| = 1}{\inf} \sum_{i=1}^{m} \| \delta^T \bU_k \|_2^2 \left(  \underset{\delta_k \in \calU_k  \atop \| \delta \| = 1}{\inf}  \| \delta_k^T \bA_k \|_2^2  \right) \:.
\end{aligned}
\label{eq:gg1}
\end{eqnarray}

Define
\begin{eqnarray}\notag
 \begin{aligned}
\calB = \min_k \left( \left\{ \underset{\delta_k \in \calU_k  \atop \| \delta \| = 1}{\inf}  \| \delta_k^T \bA_k \|_2^2  \right\}_{k=1}^{m} \right) \:. 
\end{aligned}
\end{eqnarray}
According to (\ref{eq:gg1}) and the definition of $\calB
$,
\begin{eqnarray}\notag
 \begin{aligned}
 \underset{\delta \in \calU  \atop \| \delta \| = 1}{\inf} \| \bA^T \delta \|_2^2 \ge \calB \sum_{i=1}^{m} \| \delta^T \bU_k \|_2^2 = \calB \vartheta \:.
\end{aligned}
\end{eqnarray}

Using Lemma~\ref{lm:mylemma} and the definition of $\calA$, we can bound $\calB$ as follows
\begin{eqnarray}
 \begin{aligned}
& \mathbb{P} \Bigg[ \calB < \calA \Bigg] \le \delta \:. 
\end{aligned}
\label{eq:omega_bound}
\end{eqnarray}
Therefore, according to (\ref{eq:gg1}) and (\ref{eq:omega_bound}), if the sufficient conditions of Theorem~\ref{theo:Linearly_dependant_inliers} are satisfied, then (\ref{cond:main_cond}) is satisfied and $\calU$ is recovered exactly with probability at least $1 - 3 \delta$.

\subsection{Proof of Theorem~\ref{theo:noise}}
We guarantee that a lower-bound for the Normalized Coherence Values corresponding to the inliers is larger than an upper-bound for the  Normalized Coherence Values corresponding to the outliers. First we establish the lower-bound for the   Normalized Coherence Values corresponding to the inliers. 
Note that
$$ {\bc^{*}}^T \bH {\bc^{*}} = \frac{1}{(1 + \sigma_n^2)} \|(\bA+\bE)^T \bc^{*} \|_2^2 + \|\bB^T \bc^{*} \|_2^2$$  
 where $\bc^{*}$ was defined in (\ref{eq:first11}). 
 In addition, the optimal point of (\ref{eq:first11}) when $\bd = \bd_i$ is equal to $\bc_i^{*}= \frac{(\bD\bD^T)^{-1} \bd_i}{\bd_i^T (\bD\bD^T)^{-1} \bd_i}$. According to the definition of vectors $\{ \bt_i \}_{i=n_o+1}^{M_2}$, $\|\bc_i^{*} \|_2 = \frac{\| \Sigma^{-2} \bt_i \|_2}{\bt_i^T \Sigma^{-2} \bt_i} $. When $\bd_i$ is an inlier,
$
\frac{1}{\sqrt{1+\sigma_n^2}}((\ba_i + \be_i)^T\bc^{*}) = 1 \:.
$
 Thus, 
\begin{eqnarray}\notag
 \sqrt{1 + \sigma_n^2}  + t_{\max} \sigma_n \ge \| \bU^T \bc_i^{*}\|_2 \ge \sqrt{1 + \sigma_n^2}  - t_{\max} \sigma_n 
\end{eqnarray}
 where $t_{\max}$ and $t_{\min}$ were defined in Section (\ref{sec:nooooise}). In addition, when $\bd_i$ is an inlier, $ t_{\min} \le \| \bc_i^{*} \|_2 \le t_{\max}$ and 
\begin{eqnarray*}
\begin{aligned}
 \|(\bA+\bE)^T \bc^{*} \|_2^2 & \ge \| \bA^T \bc^{*} \|_2^2 + \| \bE^T \bc^{*} \|_2^2 - 2 \sum_{i=n_o+1}^{M_2} (\ba_i^T \bc^{*})(\be_i^T \bc^{*})  \\
& \ge  \| \bA^T \bc^{*} \|_2^2 + \| \bE^T \bc^{*} \|_2^2 - 2 \sigma_n n_i t_{\max}^2. 
\end{aligned}
\end{eqnarray*}
Therefore, we can bound the Normalize Coherence Value corresponding to an inliers as follows
\begin{eqnarray}
\begin{aligned}
& \| \bD^T \bc^{*}\|_2^2 \ge \frac{(\sqrt{1 + \sigma_n^2}  - t_{\max} \sigma_n)^2}{1 + \sigma_n^2} \underset{\delta \in \calU  \atop \| \delta \|_2 = 1}{\inf} \| \delta^T \bA\|_2^2 + \\
& \frac{\sigma_n^2 \: t_{\min}^2}{1 + \sigma_n^2} \underset{ \| \delta \|_2 = 1}{\inf} \| \delta^T\bN \|_2^2 + t_{\min}^2 \underset{ \| \delta \|_2 = 1}{\inf} \| \delta^T\bB \|_2^2 - 2\sigma_n n_i t_{\max}^2 \:.
\end{aligned}
\label{eq:finalll}
\end{eqnarray}
Using (\ref{eq:finalll}) and 
 Lemma~\ref{lm:mylemma}, we can establish the flowing bound
 \begin{eqnarray*}
\begin{aligned}
& \|\bD^T \bc^{*}\|_2^2 \ge \\
&
 \frac{(\sqrt{1 + \sigma_n^2}  - t_{\max} \sigma_n)^2}{1 + \sigma_n^2} \left( \frac{n_i}{r} - \max \left( \frac{4}{3} \log \frac{2r}{\delta} , \sqrt{4 \frac{n_i}{r} \log \frac{2 r}{\delta}} \right) \right) + \\
&
\left(\frac{\sigma_n^2 \: t_{\min}^2}{1 + \sigma_2^2} + t_{\min}^2 \right)  \Bigg( \frac{n_i+n_o}{M_1} - \\
& 2\max \left( \frac{4}{3} \log \frac{2M_1}{\delta} , \sqrt{4 \frac{\max(n_i,n_o)}{M_1} \log \frac{2 r}{\delta}} \right) \Bigg) - 2 \sigma_n n_i t_{\max}^2
\end{aligned}
\end{eqnarray*}
with probability at least $1 - 3\delta$. 

Next Suppose that $\bd$ is an outlier. Similar to the proof of Theorem~\ref{theo:randomrandom}, 
define $\bd^{\perp} = \frac{\bR \bR^T \bd}{\| \bd^T \bR \|_2^2}$ where $\bR$ is an orthonormal basis for $\calU^{\perp}$. Since $\bd^T \bd^{\perp} = 1$, 
\begin{eqnarray*}
\begin{aligned}
& {\bc^{*}}^T \bH \bc^{*} \le \| \bD^T \bd^{\perp}  \|_2^2 = \| \bB^T \bd^{\perp} \|_2^2 + \frac{1}{1 + \sigma_n^2 }\| \bE^T \bd^{\perp} \|_2^2 \\
\le& \frac{1}{\| \bd^T \bR \|_2^2 } \left( \frac{n_o}{M_1} + \max \left( \frac{4}{3} \log \frac{2 M_1}{\delta} , \sqrt{4 \frac{n_o}{M_1} \log \frac{2 M_1}{\delta}}  \right) \right) +\\
& \frac{\sigma_n^2}{(1+\sigma_n^2 ) \| \bd^T \bR \|_2^2 } \Bigg( \frac{n_i}{M_1} +  \max \left( \frac{4}{3} \log \frac{2 M_1}{\delta} , \sqrt{4 \frac{n_i}{M_1} \log \frac{2 M_1}{\delta}} \right) \Bigg) \:,
\end{aligned}
\end{eqnarray*}
with probability at least $1 - 2\delta$.
According to the established lower-bound/upper-bound, if the sufficient conditions of Theorem~\ref{theo:noise} are satisfied, the Normalized Coherence Values satisfy (\ref{cond:main_cond}) with probability at least $1 - 7\delta$.

\end{document}

%% file: preamble.tex
\usepackage[mathscr]{eucal}
\usepackage[cmex10]{amsmath}
\usepackage{epsfig,epsf,psfrag}
\usepackage{amssymb,amsmath,amsthm,amsfonts,latexsym}
\usepackage{amsmath,graphicx,bm,xcolor,url}
\usepackage[caption=false]{subfig} 
\usepackage{fixltx2e}
\usepackage{array}
\usepackage{verbatim}
\usepackage{bm}
\usepackage{algorithmic}
\usepackage{algorithm}
\usepackage{verbatim}
\usepackage{textcomp}
\usepackage{mathrsfs}
\usepackage{epstopdf}

\catcode`~=11 \def\UrlSpecials{\do\~{\kern -.15em\lower .7ex\hbox{~}\kern .04em}} \catcode`~=13 

\allowdisplaybreaks[3]

\newcommand{\calA}{\mathcal{A}}
\newcommand{\calB}{\mathcal{B}}

\newcommand{\calN}{\mathcal{N}}
\newcommand{\calO}{\mathcal{O}}

\newcommand{\calU}{\mathcal{U}}

\newcommand{\ba}{\mathbf{a}}
\newcommand{\bA}{\mathbf{A}}
\newcommand{\bb}{\mathbf{b}}
\newcommand{\bB}{\mathbf{B}}
\newcommand{\bc}{\mathbf{c}}
\newcommand{\bC}{\mathbf{C}}
\newcommand{\bd}{\mathbf{d}}
\newcommand{\bD}{\mathbf{D}}
\newcommand{\be}{\mathbf{e}}
\newcommand{\bE}{\mathbf{E}}

\newcommand{\bF}{\mathbf{F}}
\newcommand{\bg}{\mathbf{g}}
\newcommand{\bG}{\mathbf{G}}

\newcommand{\bH}{\mathbf{H}}

\newcommand{\bI}{\mathbf{I}}

\newcommand{\bN}{\mathbf{N}}

\newcommand{\bq}{\mathbf{q}}
\newcommand{\bQ}{\mathbf{Q}}

\newcommand{\bR}{\mathbf{R}}
\newcommand{\bs}{\mathbf{s}}

\newcommand{\bt}{\mathbf{t}}
\newcommand{\bT}{\mathbf{T}}
\newcommand{\bu}{\mathbf{u}}
\newcommand{\bU}{\mathbf{U}}
\newcommand{\bv}{\mathbf{v}}
\newcommand{\bV}{\mathbf{V}}
\newcommand{\bw}{\mathbf{w}}

\newcommand{\bx}{\mathbf{x}}
\newcommand{\bX}{\mathbf{X}}

\newcommand{\bY}{\mathbf{Y}}
\newcommand{\bz}{\mathbf{z}}
\newcommand{\bZ}{\mathbf{Z}}

\DeclareMathAlphabet{\mathbsf}{OT1}{cmss}{bx}{n}
\DeclareMathAlphabet{\mathssf}{OT1}{cmss}{m}{sl}

\DeclareSymbolFont{bsfletters}{OT1}{cmss}{bx}{n}  
\DeclareSymbolFont{ssfletters}{OT1}{cmss}{m}{n}
\DeclareMathSymbol{\bsfGamma}{0}{bsfletters}{'000}
\DeclareMathSymbol{\ssfGamma}{0}{ssfletters}{'000}
\DeclareMathSymbol{\bsfDelta}{0}{bsfletters}{'001}
\DeclareMathSymbol{\ssfDelta}{0}{ssfletters}{'001}
\DeclareMathSymbol{\bsfTheta}{0}{bsfletters}{'002}
\DeclareMathSymbol{\ssfTheta}{0}{ssfletters}{'002}
\DeclareMathSymbol{\bsfLambda}{0}{bsfletters}{'003}
\DeclareMathSymbol{\ssfLambda}{0}{ssfletters}{'003}
\DeclareMathSymbol{\bsfXi}{0}{bsfletters}{'004}
\DeclareMathSymbol{\ssfXi}{0}{ssfletters}{'004}
\DeclareMathSymbol{\bsfPi}{0}{bsfletters}{'005}
\DeclareMathSymbol{\ssfPi}{0}{ssfletters}{'005}
\DeclareMathSymbol{\bsfSigma}{0}{bsfletters}{'006}
\DeclareMathSymbol{\ssfSigma}{0}{ssfletters}{'006}
\DeclareMathSymbol{\bsfUpsilon}{0}{bsfletters}{'007}
\DeclareMathSymbol{\ssfUpsilon}{0}{ssfletters}{'007}
\DeclareMathSymbol{\bsfPhi}{0}{bsfletters}{'010}
\DeclareMathSymbol{\ssfPhi}{0}{ssfletters}{'010}
\DeclareMathSymbol{\bsfPsi}{0}{bsfletters}{'011}
\DeclareMathSymbol{\ssfPsi}{0}{ssfletters}{'011}
\DeclareMathSymbol{\bsfOmega}{0}{bsfletters}{'012}
\DeclareMathSymbol{\ssfOmega}{0}{ssfletters}{'012}

\newtheorem{theorem}{Theorem} 
\newtheorem{lemma}[theorem]{Lemma}

\newtheorem{remark}{Remark}

\newtheorem{assumption}{Assumption}
\newtheorem{data model}{Data Model}

\newcommand{\qednew}{\nobreak \ifvmode \relax \else
      \ifdim\lastskip<1.5em \hskip-\lastskip
      \hskip1.5em plus0em minus0.5em \fi \nobreak
      \vrule height0.75em width0.5em depth0.25em\fi}